\documentclass[lettersize,journal]{IEEEtran}
\usepackage{amsmath,amsfonts}
\usepackage{algorithmic}
\usepackage{algorithm}
\usepackage{array}
\usepackage[caption=false,font=normalsize,labelfont=sf,textfont=sf]{subfig}
\usepackage{textcomp}
\usepackage{stfloats}
\usepackage{url}
\usepackage{verbatim}
\usepackage{graphicx}
\usepackage{cite}
\usepackage{tabularx} 
\usepackage{xcolor}
\usepackage{caption}
\usepackage{multirow}      % 用于合并行
\usepackage{booktabs}      % 用于专业表格线
\usepackage{makecell}      % 用于控制单元格内换行和对齐
\usepackage[table]{xcolor} % 可选，用于高亮单元格
\usepackage{siunitx}  % 对齐数字小数点
\usepackage{subcaption} % 用于子图
\usepackage{amsthm}

\theoremstyle{definition}

\theoremstyle{plain}

\begin{document}

\title{ 
 MP3: Multi-Period Pattern Pre-training for Spatio-Temporal Forecasting
% MP$^3$: Multi-Period Pattern Plugin for Spatio-Temporal Forecasting

% Decoupled multi-Period Pattern Pre-training

% STMP2: Enhancing Spatio-Temporal Forecasting
% Through Multi-Period Pattern Learning

% STPlug: A plug-and-play Multi-Period Pattern Learning Module for Traffic Flow Forecasting

% MP3: Enhancing Traffic Flow Forecasting
% through Multi-Period Pattern Learning
}

\author{
  Lilan Peng, Yandi Liu, Qingren Yao, Chongshou Li, Tianrui Li
\thanks{
 Lilan Peng, Yandi Liu, Chongshou Li, Tianrui Li are with the School of Computing and Artificial Intelligence, Southwest Jiaotong University, Chengdu 611756, China.  (E-mail: llpeng@my.swjtu.edu.cn,  LYD270816@outlook.com, lics@swjtu.edu.cn,
trli@swjtu.edu.cn)
Qingren Yao is at Eindhoven University of Technology, Eindhoven 5612 AZ, Netherlands. (E-mail:q.yao@tue.nl )
}

\thanks{Corresponding Authors: Tianrui Li (trli@swjtu.edu.cn).}
}
\markboth{Journal of \LaTeX\ Class Files,~Vol.~14, No.~8, August~2021}%
{Shell \MakeLowercase{\textit{et al.}}: A Sample Article Using IEEEtran.cls for IEEE Journals}


\maketitle

\begin{abstract}
Spatio-Temporal forecasting is crucial in diverse fields, such as transportation, climate, and energy. Urban spatio-temporal data exhibits temporal mirage: similar short-window inputs have divergent future trends, and vice versa. Existing spatio-temporal graph neural networks (STGNNs) cannot effectively identify such mirages. We argue that the core reason lies in the short-window inputs that have incomplete period observation, heterogeneous global spatial correlation, and
cross-period superposition causality. 
% an inherent scale limitation of short-window observations, which perceive only short-scale period patterns within the narrow window, but miss the long-scale period patterns that embed each fragment within the full temporal structure. This limitation manifests in 
% three critical dimensions: 
% We argue that the core reason is that two short time series with the same trend belong to different period patterns in a complete long time series, have heterogeneous global spatial correlation, and cross-period superposition causality. 
To bridge this gap, we develop a novel \underline{M}ulti- \underline{P}eriod  \underline{P}attern  \underline{P}re-training (MP3), a plug-and-play pre-training plugin for distinguishing temporal mirages.  
MP3 presents two core innovations: 
(1) The multi-period pattern learning is designed to learn multi-period patterns from long time series. Specifically, multi-period temporal modeling leverages edge convolution to identify different multi-period patterns. Multi-period spatial modeling uses a bottleneck project and a global memory bank to capture heterogeneous global spatial relations efficiently. Cross-period pattern interaction employs a causality-enhanced Transformer to capture dependencies across
different period patterns. 
(2) This plugin can seamlessly integrate into existing STGNN backbones to strengthen their forecasting performance. 
The experiment on five STGNN baselines across five real-world datasets (including a large-scale dataset CA) verify the effectiveness, superior scalability and strong adaptability of MP3, which brings consistent and robust performance improvements across all evaluated baselines. On average, MP3 reduces the MAE 4.7\% and the RMSE 5.0\%.
The code can be available at https://github.com/YAN-outlook/MP3.

\end{abstract}

\begin{IEEEkeywords}
 Spatio-Temporal Graph Neural Network, Pre-Training Plugin, Multi-Period Patterns Learning.
\end{IEEEkeywords}

\section{Introduction}
% As smart cities and intelligent transportation systems advance, the transportation field is modernizing to handle pressing problems like traffic congestion and pollution resulting from urban expansion . Traffic flow forecasting leverages historical observation data to predict future conditions, supporting sustainable mobility, traffic signal optimization, route planning, and shared-vehicle scheduling, thus empowering modern urban transportation systems with greater efficiency and intelligence.\cite{liu2025review,ali2024survey}.

Spatio-Temporal forecasting refers to analyzing historical observation data to reveal potential evolution patterns of spatio-temporal data and predict future conditions. It has attracted widespread attention due to its critical impact in diverse real-world applications, including intelligent transportation \cite{sayed2023traffic}, crime prediction \cite{Sec4li2022spatial}, and air quality prediction\cite{Sec1hettige2024airphynet}. 
Early work in spatio-temporal prediction relied primarily on statistical models (e.g., Historical Average (HA)\cite{Smith_Demetsky_1997} and ARIMA \cite{Anderson_Box_Jenkins_1978}, \cite{Lippi_Bertini_Frasconi_2013}) and early machine learning techniques such as vector Auto-Regression \cite{Lütkepohl_2005}, \cite{Zivot_Wang_2003}. 
% A key limitation of these methods is their difficulty in modeling the complex nonlinear interactions within large-scale traffic networks for traffic flow prediction. 
The advent of spatio-temporal big data has prompted a paradigm shift towards data-driven deep learning approaches\cite{zhang2017deep,Ma_Tao_Wang_Yu_Wang_2015, Hochreiter_Schmidhuber_1997, Zhang_Zheng_Qi_Li_Yi_2016}, which excel at learning the inherent spatio-temporal dependencies in dynamic systems\cite{Lv_Duan_Kang_Li_Wang_2014}. 
\begin{figure} %  
  \includegraphics[width=1\linewidth]{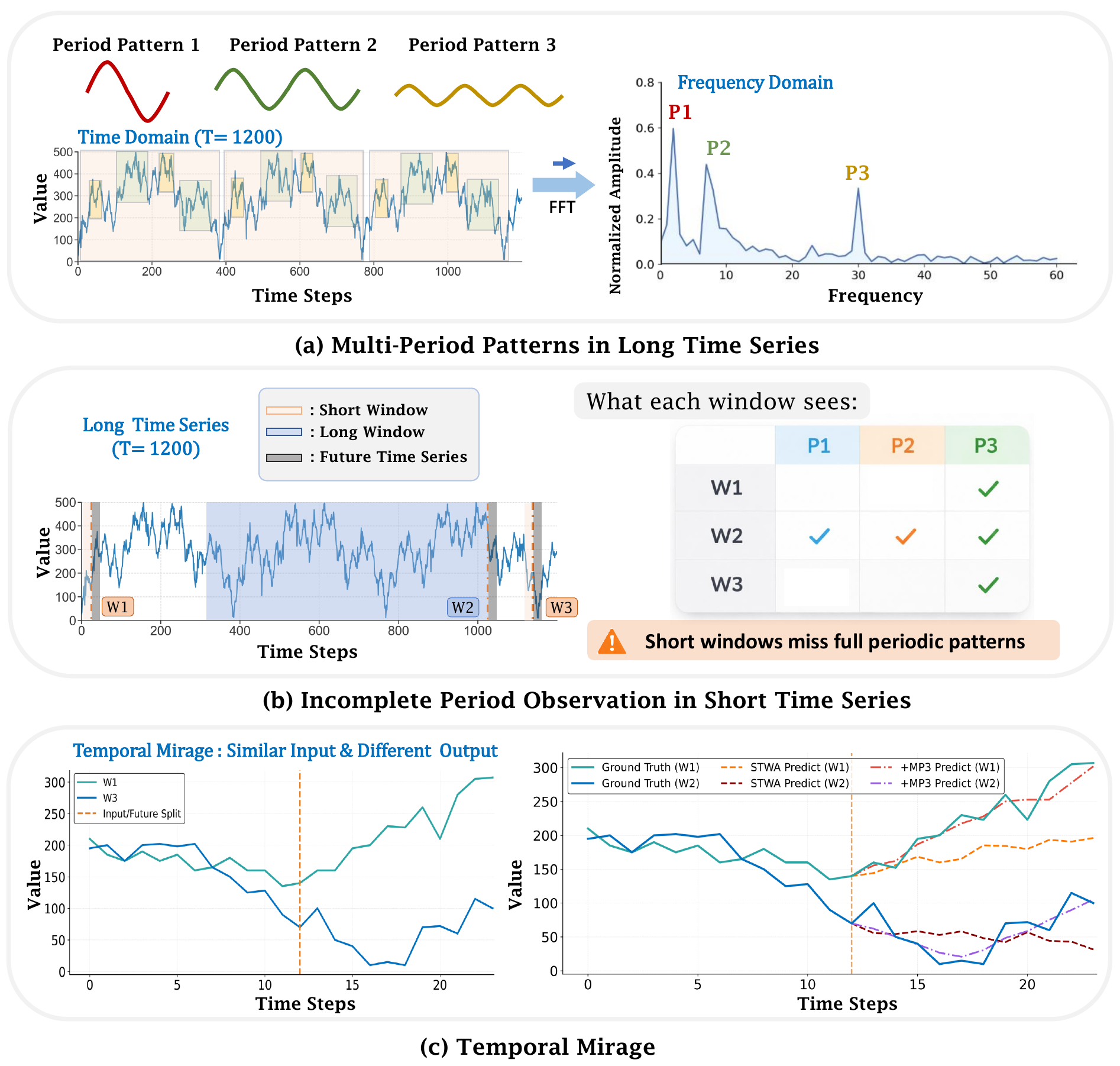} 
\caption{Figure 1 \textbf{Period Patterns for Long and Short Time Series and Temporal Mirage.} As shown in Figure (a), long time series exhibit rich period patterns (period pattern 1, period pattern 2, period pattern 3), which can be clearly reflected in the frequency domain through dominant peaks (P1, P2, P3). In contrast, Figure (b) shows that short windows capture only partial period information, failing to identify temporal mirages: similar inputs (W1, W3) have divergent future trends, as depicted in Figure (c). Our plugin MP3 is capable of effectively distinguishing temporal mirages by learning multi-period patterns from long time series.}
  \label{fig1:Motivation}
\end{figure}
Recent significant efforts  \cite{jin2023spatio} have been devoted to designing increasingly intricate spatio-temporal graph neural networks (STGNNs), such as improved graph convolution networks \cite{li2017diffusion, Zhao_Song2020, Fang_Long_Song_Xie_2021, Ye_Sun_Du_Fu_Xiong_2022, Wang_Zhang_Wang_Chen_2023}, adaptive graph structure learning \cite{Zhang_Chang_Meng_Xiang_Pan_2020,ye2022learning, Mamb2024,bai2020adptive,GraphwaveNet_2019,wu2020connecting,jiang2023spatio,dong2024heterogeneity}, efficient attention mechanisms\cite{zheng2020gman,SSTBAN_Guo_2023,PDFormer2023,ASTGCN_Guo_2022,liang2023airformer}, and other advanced frameworks\cite{cao2025spatiotemporal,Pan_Liang_Wang_Yu_Zheng_Zhang_2019,FastST2024,zhou2023maintaining}, which yield incremental yet meaningful performance gains. 
% Nevertheless, 

Despite continuous architectural innovations, performance improvements have gradually slowed and tend to saturate, which can be primarily attributed to a key problem: 
% \textbf{incomplete period observations in short time series hinder the identification of temporal mirage}.
most of these existing models struggle to identify temporal mirages. 
As shown on the left in Figure 1(c), spatio-temporal data exhibit a \textit{temporal mirage} phenomenon, where short-term inputs can be very similar, yet the future trends are completely different; conversely, similar future trends may be from entirely identical short-term inputs. We argue that the reasons arise from three main factors when observations are confined to short windows: \textit{incomplete period observation, heterogeneous global spatial correlation, and cross-period superposition causality}. 

\textit{Incomplete period observation:} Most current models usually use a small window input, e.g., the past twelve time steps (one hour) to predict the future twelve time steps, which is blind to context information (e.g., multi-period patterns) beyond the window.
As shown in Figure 1(b), short windows merely capture local, incomplete period observation. 
On the contrary, long time series, as illustrated in Figure 1(a), inherently contain diverse multi-period patterns (e.g., period pattern 1, 2, and 3), which can be clearly presented as distinct prominent peaks (P1, P2, and P3) in the frequency domain. Moreover, two similar short-term trends may exist in a completely different position of a long time series. For example, the local pattern of 'a continuous increase in passenger flow over one hour' could occur during the 'initial phase of the morning peak' (suggesting continued future growth) or during the 'peak-to-decline phase of the morning peak' (suggesting an impending drop). Short time series fail to encode the absolute period position to which they belong, leading to misjudgments about the future. 

\textit{Heterogeneous global spatial correlation:} Similar short-term temporal dynamics may correspond to completely different global spatial association patterns. 
For instance, the same local pattern of 'increasing morning peak flow' can appear in a 'commuting core business district' versus a 'suburban scenic area'. The underlying global spatial association (such as origin-destination patterns of passenger flow) is entirely different in these two spatial contexts. Short-window inputs lack sufficient statistical evidence to distinguish these context-dependent spatial patterns, leading models to make erroneous inferences.

\textit{Cross-period superposition causality:} We observe that multi-period patterns exhibit unidirectional causal influence from strong to weak period patterns. This inherent causal direction is determined by the strength of each pattern, not its cycle length. For instance, taking 'passenger flow at 8:00 AM' as an example, it depends not only on the immediate trend of the 'past five minutes' but is also jointly shaped by long-cycle patterns such as 'whether it is the morning peak (intraday period)' and 'whether it is a working day (weekly period)'. On workdays, the intraday morning peak is a strong pattern and exerts one-way causal effects on short-term fluctuations. On weekends, the weekly pattern becomes dominant, unidirectionally suppressing the usual morning peak.

Fundamentally, all three challenges stem from the inability of short-window inputs to support explicit modeling of multi-period patterns in long spatio-temporal sequences. An intuitive solution to the above challenges is to adopt an end-to-end model. However, existing models rely heavily on intricate and tailored designs; any attempt to decouple or reassemble their components often results in degraded prediction performance, lacking scalability and generality. Moreover, the computational and memory cost of STGNNs grows exponentially when the input window length increases.
Pre-training strategies have achieved remarkable success in natural language processing \cite{devlin2019bert} and computer vision\cite{he2022masked}, and have recently inspired initial explorations in spatio-temporal forecasting\cite{shao2022pre,li2023gptst,gao2024spatial}. By learning context-rich representations, pre-training can effectively enhance various downstream models. 
Motivated by these advantages, we propose a scalable and general multi-period pattern pre-training framework \textbf{MP3} that can be seamlessly integrated into nearly arbitrary STGNNs and applied across a wide range of spatio-temporal datasets. The core of MP3 is a multi-period pattern learning module, which systematically solves the three aforementioned challenges.
Multi-period temporal modeling employs edge convolution to decouple independent patterns of different periods from long time series and learn intra- and inter-period temporal variation dependencies, thereby obtaining a unique multi-period position identity to resolve the \textit{incomplete period observation} problem.
Multi-period spatial modeling represents large-scale, heterogeneous, and global spatial interactions through bottleneck projection and global memory bank strategies, directly capturing the \textit{heterogeneous global spatial correlation}.
Cross-period pattern interaction leverages a causal-enhanced transformer to learn interactions across various period patterns, explicitly learning the \textit{cross-period superposition causality} that is invisible in short windows.
% To solve this challenge, we develop MP3, a plug-and-play pre-training plugin designed to capture complex multi-period patterns from long-term historical spatio-temporal sequences and enhance the performance of existing STGNNs. The core of MP3 is a multi-period pattern learning module with three components. Specifically, multi-period temporal modeling captures intra- and inter-period temporal dependencies, multi-period spatial modeling models node-wise spatial interactions, and cross-period pattern interaction explores complex correlations among different Period patterns. 
% Notably, MP3 is a general plugin that can seamlessly integrate into almost arbitrary STGNNs and a wide range of spatio-temporal datasets, from small-scale to large-scale.
Our contributions are delivered as follows.
\begin{itemize}
    \item We introduce MP3, a scalable and general plug-and-play pre-training module to capture multi-period patterns from long-term spatio-temporal data. Built upon multi-period temporal modeling, multi-period spatial modeling, and cross-period pattern interaction, MP3 effectively models spatio-temporal dependencies in a period pattern, as well as causal correlations across different period patterns.
    \item We seamlessly integrate MP3 into existing STGNN backbones, significantly enhancing forecasting performance. Specifically, we freeze the parameters of the pre-trained MP3 and employ a gating mechanism to adaptively fuse the extracted multi-period patterns with short-term inputs.
    % , thereby enhancing the forecasting performance of existing STGNNs.
    \item  We evaluate MP3 on four publicly available spatio-temporal datasets and a large-scale dataset CA with five representative STGNN backbones. The experimental results demonstrate that MP3 consistently improves prediction performance on different models and datasets, confirming its stronger capability in modeling multi-period patterns from long-term spatio-temporal data. Furthermore, notable gains on the large-scale CA dataset demonstrate MP3's strong scalability. 
\end{itemize}

\section{Related Work}
\subsection{Spatio-Temporal Prediction}
Spatio-Temporal forecasting, which aims to employ historical observation data to predict future conditions, has long been a focus of research\cite{wang2021libcity, TVT2025}. Early statistical approaches gained some improvement but frequently proved inadequate in capturing dynamic spatial correlation and complex spatio-temporal dependencies in spatio-temporal data\cite{jin2023spatio}. To solve these issues, attention has shifted toward deep learning frameworks, which excel at learning complex and dynamic features, like nonlinear spatio-temporal correlations\cite{li2017diffusion,yu2018spatio}.

STGNNs\cite{jin2023spatio} bring a paradigm shift in spatio-temporal prediction by seamlessly integrating graph neural networks\cite{kipf2017semi} and temporal modeling methods\cite{Zhao_Song2020}. Several influential STGNNs \cite{li2017diffusion, GraphwaveNet_2019, yu2018spatio, bai2020adptive} have been developed, further enhanced by attention mechanisms \cite{zheng2020gman} for dynamic relation modeling.
% However, recent progress in STGNN design has led to marginal improvements. Therefore, the field is witnessing a new paradigm shift, with researchers increasingly integrating Large Language Models (LLMs) \cite{huang2024std,liu2025stllm_plus}  to unlock further advances.
Several advanced methods have focused explicitly on modeling spatio-temporal heterogeneity. Specifically, Gao et al. \cite{gao2024spatial} devise a decoupled masked strategy to distill spatial and temporal knowledge in the pre-training stage \cite{gao2024spatial}. A heterogeneity-informed learning Network \cite{dong2024heterogeneity} leverages dedicated spatial, temporal, and spatio-temporal meta parameter learning to model complex and dynamic spatio-temporal heterogeneity. Furthermore, self-supervised learning paradigms for multi-modal forecasting integrate contrastive learning and data augmentation to capture cross-modal heterogeneous patterns\cite{ijcai2024p223}.
Nevertheless, these methods suffer from a critical limitation: they cannot provide representations for multi-period patterns in long-range traffic sequences. Existing pre-training plugins only focus on generic spatio-temporal knowledge distillation, while neglecting the inherent period that dominates spatio-temporal variations. To fill this gap, we design a plug-and-play pre-training plugin MP3, which explicitly models and distills multi-period pattern knowledge to enhance spatio-temporal forecasting.

\subsection{Period Patterns Modeling}
Modeling Period patterns is critical in spatio-temporal prediction. Early research employed various methods to capture these patterns: Zhang et al. 
\cite{zhang2017deep} utilize convolutional neural networks to separately capture daily and weekly cycles in traffic data; Ye et al. \cite{ye2022learning} analyze the evolutionary patterns of graph structures across different temporal scales through temporal feature aggregation; Ma et al. \cite{Mamb2024} design a time-aware graph learning method, combining temporal difference learning with a period discriminant function to construct dynamic spatio-temporal graphs, thereby capturing spatial dependencies with both trend and period features. However, these multi-period modeling approaches generally rely on short-term time series for period pattern identification, failing to adequately learn the complex and rich period regularities present in long-term time series.

For time series prediction, TimesNet \cite{wu2023timesnet} reshapes 1D time series into 2D tensors, using a CNN-based hierarchical structure to learn correlations across different time scales. However, directly employing convolutional operations to extract multi-period features may lead to information loss and reduced accuracy. Furthermore, although TimesNet presents 2D reshape visualizations, explaining their correspondence to the original 1D temporal semantics remains challenging. MSGNet\cite{MSGNet2024AAAI} employs a graph convolution module to capture inter-series relations and utilizes an attention mechanism to model intra-series dependencies, yet its adaptive graph convolution mechanism significantly increases model complexity and computational overhead. MICN\cite{micn2023} integrates local feature extraction and global correlation modeling through a multi-scale convolutional architecture, but its performance is highly sensitive to hyperparameter tuning (e.g., convolution scales), and its fully convolutional design sacrifices model interpretability.

To model complex temporal dynamics in long sequences, recent work leverages transformer-based encoders within pre-training modules, thereby extracting robust long-term dependencies. For instance, STD-MAE \cite{gao2024spatial} devises a decoupled masking strategy across spatial and temporal dimensions to distill spatio-temporal knowledge in pre-training stage. Han et al. \cite{BigST2024} develop a linearized Transformer to learn long-term time series features in pre-training stage. While these approaches can represent long-term spatio-temporal dependencies, they fall short in distilling multi-period knowledge. Therefore, we provide a novel plug-and-play pre-training plugin MP3 that extracts multi-period pattern knowledge
from long-term sequences and ensembles them into STGNN backbones to strengthen the performance of spatio-temporal prediction.

\begin{figure*}[!htb] 
  \centering
  \includegraphics[width=\textwidth]{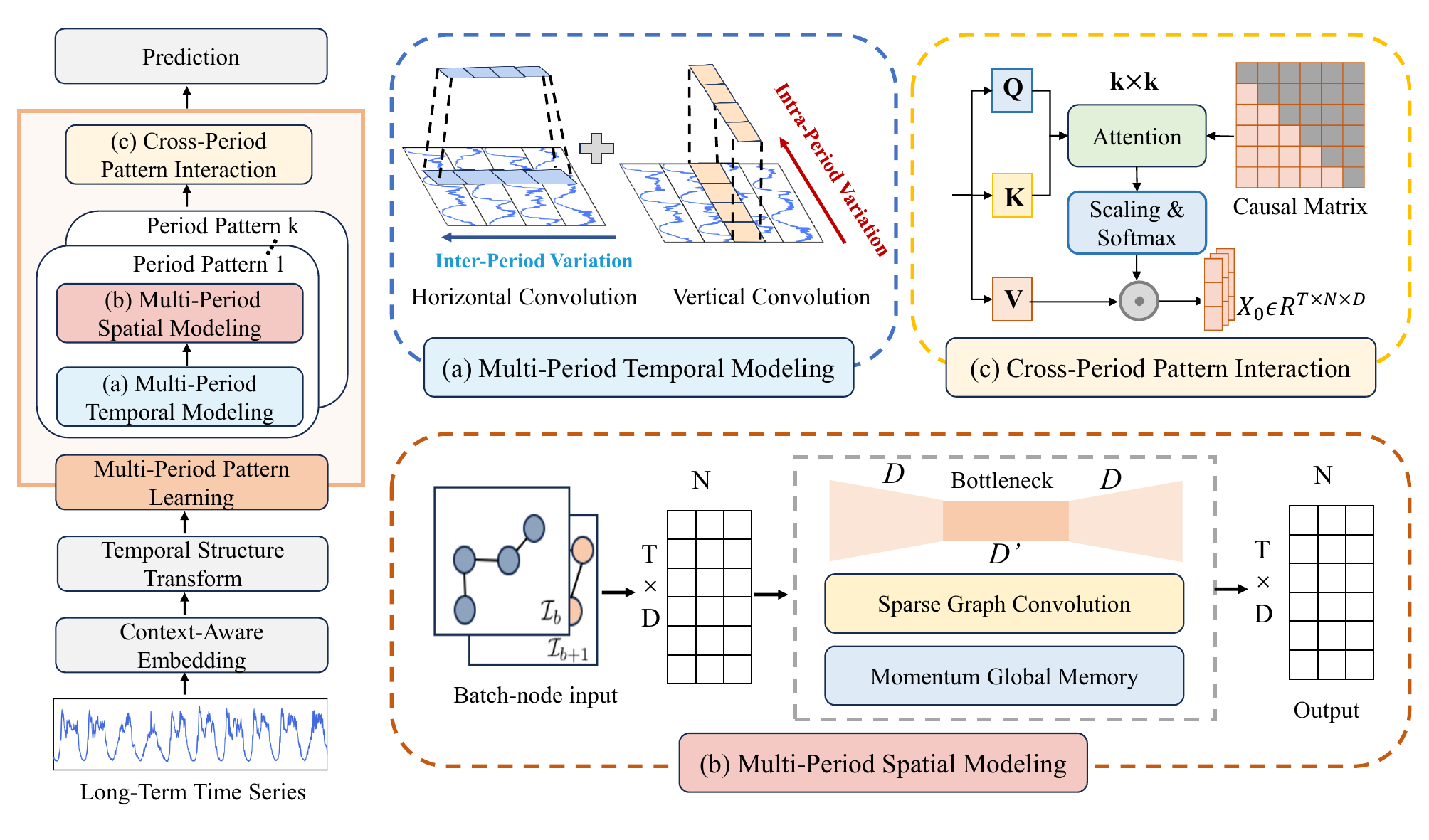}  
\caption*{Figure 2 depicts the framework of MP3. 
The core module is multi-period pattern learning, which integrates multi-period temporal modeling, multi-period spatial modeling, and cross-period pattern interaction to capture multi-period patterns in long-term time series. Multi-period temporal modeling captures both intra- and inter-period variations via edge convolution. Multi-period spatial modeling then efficiently models heterogeneous global spatial correlation using bottleneck projection and memory bank strategies. Finally, cross-period pattern interaction leverages a causality-enhanced Transformer to capture dependencies across different period patterns.
}
  \label{fig2:main}
\end{figure*}

\section{METHODOLOGY}
\textbf{ Spatio-Temporal Forecasting: } 
Given historical T-step spatio-temporal data  $\mathbf{X}_{t-T+1:t}$ and graph network G,
spatio-temporal forecasting for the next $Q$ time steps can be defined as follows: 

\begin{equation}
\label{eq:forecasting_problem}
\hat{\mathbf{X}}_{t+1:t+Q} = F_\theta \left( \mathbf{X}_{t-T+1:t}; G \right)
\end{equation}

Here \( F_\theta \) represents the objective function, and \( \mathbf{X}_t = (x_t^1, x_t^2, \dots, x_t^N) \in \mathbb{R}^{N\times D} \) is real observation values of $N$ nodes at time t with \textit{D} input dimensions. 

MP3 is a plug-and-play pre-training plugin, as is depicted in Figure 2, that extracts and distills multi-period pattern knowledge from long-term time series. By seamlessly integrating into existing STGNN backbones, MP3 significantly enhances its ability to model complex long-term multi-period spatio-temporal dependencies during the forecasting stage. First, we perform context-aware embedding (section \ref{secta}) on long-term spatio-temporal data, followed by a temporal structure transform (section \ref{sectb}) that extracts diverse period features and transforms 1D time series into a 2D representation. Furthermore, we design a multi-period pattern learning module, which consists of three components: multi-period temporal modeling (section \ref{sectc}), multi-period spatial modeling (section \ref{sectd}), and cross-period pattern interaction  (section \ref{secte}). Multi-period temporal modeling captures dependencies of intra and inter-period variations via edge convolution. Multi-period spatial modeling efficiently learns multi-period spatial dependencies using bottleneck projection and memory bank strategies. The cross-period pattern interaction employs a causality-enhanced Transformer to capture dependencies among different period patterns.
% (2) Downstream Traffic Flow Prediction: A gating mechanism fuses pre-trained distilled Period features and short-term traffic flow data, then feeds the fused features into the backbones (existing traffic flow prediction model) to generate final outputs. 

% a) \textbf{Multi-period temporal modeling}: Implemented through edge convolution for efficiency;

% b) \textbf{Multi-period spatial modeling}: Uses memory bank aided GCN to fix adjacency disruption, which arises when large urban networks input nodes in batches;

% c) \textbf{Cross-period pattern attention}: Achieved through an upper-triangular attention matrix via cross-period pattern attention. 

\subsection{Context-Aware Embedding}\label{secta}
Formally, we start with the historical observation sequence $ X_{t-T+1:t}$.
The context-aware embedding (CAE) module is devised to preserve the intrinsic period feature of spatio-temporal data while compressing long input sequences to reduce computational complexity. Crucially, leveraging convolutional translation invariance, CAE aligns distant recurring patterns, thereby preserving long-range inter-period dependencies.
% The CAE module relies on two key steps to optimize efficiency and enhance context correlation: (a) Context Convolution for Temporal Downsampling, and (b) Learnable Temporal Positional Encoding.
% \paragraph{Context Convolution for Temporal Downsampling}
We first apply a context convolutional layer to the input tensor $X_{t-T+1} \in \mathbb{R}^{T \times N \times D}$. 
% The convolution operation is formulated as:
% \begin{equation}
% \boldsymbol{X}_{\text{CAE}} = \boldsymbol{X} \star f(t) = \sum_{m=0}^{M-1} f(m) \cdot \boldsymbol{X}_{t-\omega \times m}
% \end{equation}
% Here  $ \omega $  denotes the dilation factor and  $ f(\cdot) $  is a filter kernel of length  $ M $.
This operation compresses the temporal length of the sequence while preserving core period patterns like daily morning and evening peaks.

% \paragraph{Learnable Temporal Positional Encoding}
% To preserve the absolute temporal position information of downsampled sequences, we introduce a learnable positional embedding matrix $\boldsymbol{P}_e \in \mathbb{R}^{L\times D}$, where $L$ and $D$ denote the sequence length and feature dimension, respectively. The positional embedding is fused with the convolutionally encoded features via element-wise addition:
% \begin{equation}
% \boldsymbol{X}_{\text{E}} = \boldsymbol{X}_{\text{CAE}} + \boldsymbol{P_e}
% \end{equation}

% This operation retains the inherent temporal order of the time series, and provides consistent position-aware representations for modeling Period dependencies.

\subsection{Temporal Structure Transform}\label{sectb}
 % To solve two persistent shortcomings of existing methods: (1) the use of rigid, explicit period patterns (such as Weekly, Daily) and (2) a constrained focus on short input sequences. 
 Each time point exhibits two types of temporal variations: (1) intra-period variation among adjacent time points within the same period, and (2) inter-period variation across different periods at the same phase. For instance, in a one-week cycle, t=0 days and t=7 days (i.e., next Monday) represent distinct temporal instances yet share the same phase, as both align with the early morning start of Monday. However, conventional 1D temporal representations are limited to capturing only intra-period variations between adjacent time points, failing to model inter-period dependencies that share the same phase. To tackle this issue, we design a temporal structure transform module that includes multi-period identification, main period selection, and time series reshaping to accurately mine and structurally represent dynamic implicit periods in long-term time series. 
 % The implementation details for each module are described below. 

\paragraph{Multi-Period Identification} The insight of this module is to extract multi-period features via frequency-domain analysis.
Inspired by TimesNet \cite{wu2023timesnet}, we extract the dominant period components from the embedded sequence $X_{\text{E}}$ using the Fast Fourier Transform (FFT). 
\begin{equation}
    F = \text{Avg}\big(\text{Amp}(\text{FFT}(X_{\text{E}}))\big)
\end{equation}

Specifically, we first apply $\text{FFT}(\cdot)$ to transform the embedded sequence into frequency domain, then use the $\text{Amp}(\cdot)$ operator to extract the amplitude values corresponding to each frequency, and averaged over the feature dimension \textit{D} via $\text{Avg}(\cdot)$ to obtain the aggregated frequency vector $F \in \mathbb{R}^{L}$, where $L$ denotes the frequency dimension. Thus, Vector F provides a frequency-domain feature of ${X}_{\text{E}}$.

% 图片下方直接接双栏文本，无间距也可
\paragraph{Main Period Selection} \label{main-period-selection}
Temporal dynamics often exhibit non-stationary period features, and the frequency-driven period selection mechanism enables the model to track time-varying period scales adaptively. Therefore, based on the aggregated frequency $F$, we further select the set of main period lengths $\{p_1, \dots, p_k\}$. Specifically, we apply the $\arg\text{Top-k}$ function to identify the k largest amplitudes corresponding to the top-\textbf{k} strongest period patterns, obtaining the dominant frequencies $\{f_1, \dots, f_k\}$ associated with the corresponding period patterns $\{p_1, \dots, p_k\}$sorted in descending order of amplitude strength. This ordering defines the causal interaction sequence in Section E. The formulation is as follows:
% The hyperparameter 
% $k$ controls the number of Period scales considered in the following analysis.

\begin{equation}
    \begin{aligned}
    f_1, \ldots, f_k = 
    \operatorname*{\arg \text{Top-k}}_{f_* \in \{1, \ldots, L/2\}}(F) \\
    p_i = \frac{L}{f_i}.
    \end{aligned}
    \end{equation}
% \paragraph{ Multi-Scale Period Tensor Reconstruction: Structured Representation of Period Patterns}
\paragraph{Time Series Reshape}
 We argue that although the strength of period patterns may fluctuate locally, the structural dependencies across different period patterns remain relatively stable, which allows the model to learn consistent intra-series and inter-series correlations. Therefore, after determining the main period patterns $\{p_1, \dots, p_k\}$ and their corresponding frequencies $\{f_1, \dots, f_k\}$, we explicitly decouple period correlations through time series reconstruction. For each period pattern $p_i$, we first perform zero-padding on the embedded sequence $X_{\text{E}}$ to ensure compatibility with the $Reshape_{p_i, f_i}$ operation, and convert the 1D time series $X_E$ into a set of 2D tensors $X_i \in \mathbb{R}^{N \times p_i \times f_i \times D}$ through the following transformation:
\begin{equation}
X_i = \text{Reshape}_{p_i, f_i}(\text{Padding}(X_{\text{E}})), \quad i \in \{1, \dots, k\}
\end{equation}

Through this transformation, the original sequence is reorganized according to different period patterns. This not only retains intra-period variations within individual period tensors, but also establishes inter-period variations across different period slices.
% \begin{figure*} %  
%  \centering
%   \includegraphics[width=\linewidth]{EdgeConv.pdf} 
%     \caption{Figure 5 \textbf{Comparison of 2D Convolution (Left) and Edge Convolution (Right).}
%     Different from 2D convolution, the edge convolution leverages the factorized convolutions (horizontal convolution and vertical convolution) to capture the correlation of inter-period and intra-period variation. This novel convolution has a low parameter count and is a lightweight method.
%     }
%   \label{fig1:Motivation}
% \end{figure*}

\begin{figure} %  
 \centering
  \includegraphics[width=1\linewidth]{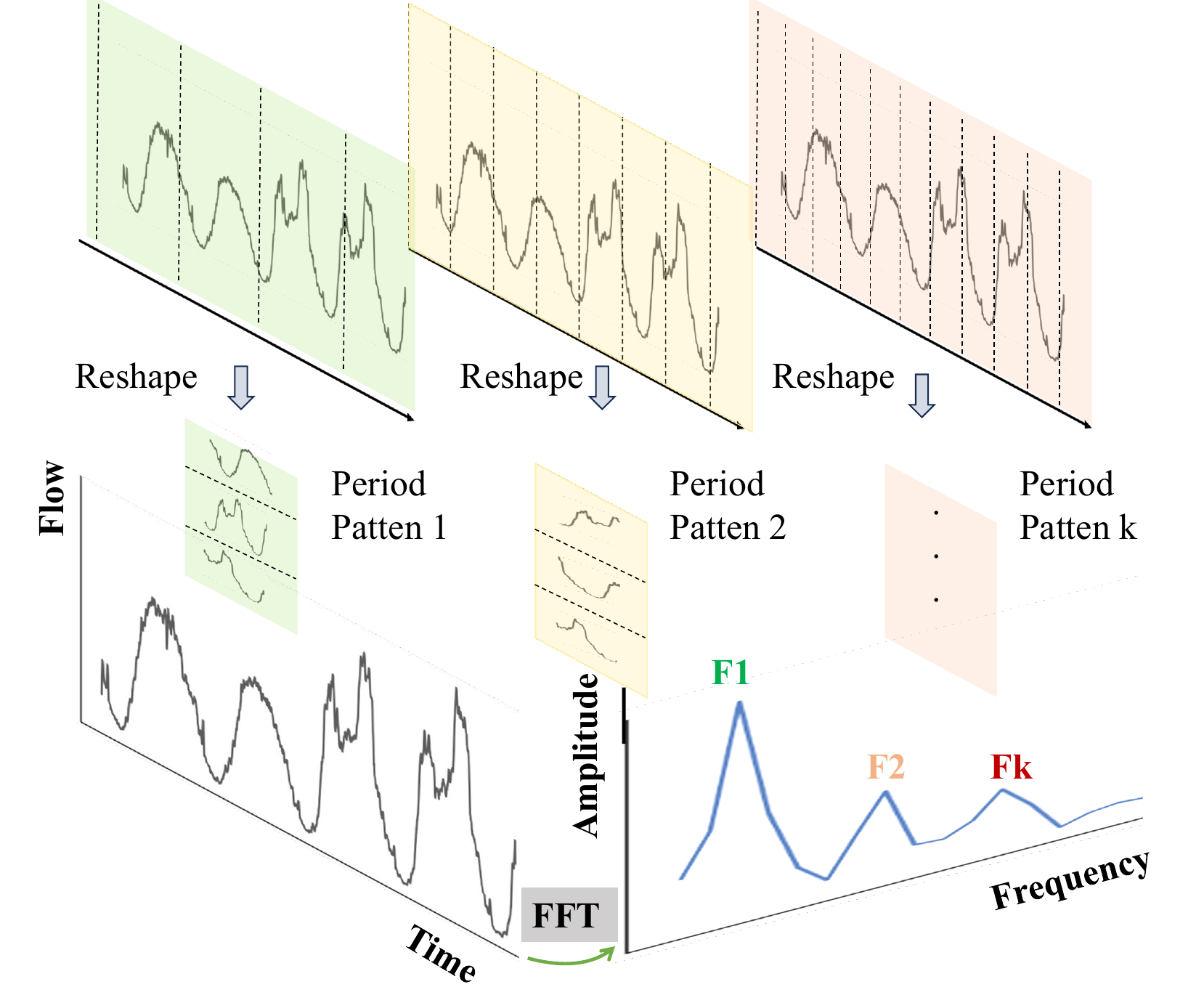} 
    \caption{Figure 3 \textbf{2D Time Structures.} We leverage Fast Fourier Transform to transform 1D time series into 2D time variances and select the top-k prominent period patterns. }
  \label{fig3:Motivation}
\end{figure}

\subsection{Multi-Period Temporal Modeling} \label{sectc}
To tackle the inability of short-window inputs to derive discriminative multi-period positional representations, we need to decouple and model intra-period and inter-period temporal variations separately from long time series.
We observe that intra-period variations and inter-period variations are relatively independent in temporal dependencies, which is also validated in section \ref{2Dstructure}. Thus, different from the 2D-Inception Block leveraged in TimesNet \cite{wu2023timesnet}, we devise a multi-period temporal modeling block to model intra-period and inter-period variations separately,  achieving dense and decoupled feature learning for both temporal variations. Specifically, we adopt edge convolution for temporal modeling, which comprises two parallel 
operations: horizontal convolution $C_h(X_i)$ with an inception-inspired set of  $[1,2l+3]$ kernels, where $l=0,1,\dots, P-1$, operating along the $f_i$ dimension to capture inter-period variations, and P denotes the set number of convolution kernels. Meanwhile, the vertical convolution $C_v(X_i)$ employs a corresponding set of $[2l+3,1]$ kernels on the $p_i$ dimension to capture intra-period variations. The formulation is as follows:

\begin{equation}
\begin{split}
C_{h}(X_i) &= \sum_{l=0}^{P-1} \text{Conv}(X_i, K=(1, 2l+3)) \\
C_{v}(X_i) &= \sum_{l=0}^{P-1} \text{Conv}(X_i, K=(2l+3, 1))
\end{split}
\end{equation}

% 第二步：主公式中引用已定义的卷积简写
\begin{equation}
X_{i}^{\text{Edge}}= C_{h}(X_i) + C_{v}(X_i)
\end{equation}

Subsequently, an adaptive average pooling layer~\cite{Deeper2015} is applied to resolve scale discrepancies induced by the convolutions. Crucially, this operation restores the feature map dimensions to $(p_i, f_i)$, ensuring that the total spatial size strictly matches the original time series length (i.e., $p_i \times f_i = T$). Finally, a reshape operation projects the 2D time series back to the 1D time series:

\begin{equation}
    X_{i}^t = \text{Reshape}\Big( \text{AdaptivePool}\big( X_{i}^{\text{Edge}}, \, (p_i, f_i) \big) \Big) \in \mathbb{R}^{N \times T \times D}
    \label{eq:reshape_output}
\end{equation}

\subsection{Multi-Period Spatial Modeling}\label{sectd}
To model heterogeneous global spatial correlation, global topological perception is required to distinguish context-dependent spatial patterns. However, this process faces three practical constraints: prohibitive computational cost for large node sets, global structure loss due to batch sampling, and spatial feature redundancy. To alleviate the above issues, we devise 
% a memory bank-aided multi-period spatial modeling module, which consists of three main components: 
a spatio-temporal bottleneck projection for dimension reduction, a sparse graph convolution on dynamic sub-graphs, and a momentum memory bank for global context restoration.

Formally, let the input batch be $\mathbf{X_i^t} \in \mathbb{R}^{N_b \times T \times D}$, where $N_b$ is the number of sampled nodes indexed by $\mathcal{I}_b$. The block transforms the input into the output $\mathbf{X_i^s}$ via a unified residual mapping:

\begin{equation}
X_i^s = \mathcal{P}_{\text{out}} \left( \mathcal{N} \Big[ \underbrace{\mathcal{G}_{\theta} \big( \Phi(\mathbf{X_i^t}) \big)}_{\text{Local Aggregation}} + \underbrace{\mathcal{E} \big( \mathcal{M}[\mathcal{I}_b] \big)}_{\text{Global Memory}} \Big] \right)
\end{equation}

where $\mathcal{N}(\cdot)$ represents the layer normalization operation; $\mathcal{E}(\cdot)$ is the expansion operator that aligns dimensions via broadcasting; and $\mathcal{P}_{\text{out}}$ denotes the linear projection to the output dimension.
The process involves three steps:

\paragraph{Spatio-Temporal Bottleneck Projection}
We first reduce the feature dimension from $D$ to $D' = D/4$ and extract local patterns using a 2D convolution with kernel size $3 \times 3$:
\begin{equation}
\mathbf{H}=\Phi(X_i^t) = \sigma \Big( \text{BN} \big(X_i^t \ast \mathbf{W}_{\phi} \big) \Big)
\end{equation}

where $\ast$ denotes convolution on period and feature dimensions,  $\mathbf{W}_{\phi}$ denotes the learnable convolutional kernel parameters, $\sigma$ is LeakyReLU, and BN is batch normalization. This step removes redundant information while preserving key spatial correlations.

\paragraph{Sparse Graph Convolution}  
Next, we aggregate local spatial features using a sub-adjacency matrix $\tilde{\mathbf{A}}_b$, which is dynamically extracted from the global adjacency matrix $\mathbf{A}$ based on the current node indices $\mathcal{I}_b$. The operation yields the local feature representation $\mathbf{Z}_{local} \in \mathbb{R}^{N_b \times T \times D'}$:
\begin{equation}
\mathbf{Z}_{local} = \mathcal{G}_{\theta}(\mathbf{H}) = \mathcal{N} \Big( \big( \tilde{\mathbf{A}}_b \otimes \mathbf{H} \big) \mathbf{W}_{\theta} \Big)
\end{equation}

Here, $\otimes$ represents the sparse spatial aggregation operator, which implicitly aligns dimensions to perform message passing between neighboring nodes. $\mathbf{W}_{\theta}$ is a learnable projection matrix, and $\mathcal{N}(\cdot)$ denotes Layer Normalization. This allows efficient modeling of local dependencies without processing the full graph.

\paragraph{Momentum Global Memory}  
To recover global structural information lost during batch sampling, we maintain a global memory bank $\mathcal{M}[\mathcal{I}_b] \in \mathbb{R}^{N \times D'}$. It is updated at each node-batch using an Exponential Moving Average (EMA) of the batch's static spatial features $\bar{\mathbf{z}} \in \mathbb{R}^{N_b \times D'}$:
\begin{equation}
\mathcal{M}[\mathcal{I}_{b+1}] \leftarrow \mu \cdot \mathcal{M}[\mathcal{I}_b] + (1 - \mu) \cdot \left( \frac{1}{T} \sum_{t=1}^{T} \mathbf{Z}_{\text{local}}^{(t)} \right)
\end{equation}

where $\mathcal{M}[\mathcal{I}_b]$ represents the memory vectors associated with the current batch indices $\mathcal{I}_b$, and $\mathcal{I}_{b+1}$ denotes the next node-input batch.
The retrieved memory vectors are broadcast across the time and period dimensions to match the shape of local features, and then added directly. This effectively supplements the current batch with global structural information that would otherwise be missing due to node sampling. Finally, $\mathcal{P}_{\text{out}}$ projects the fused features to the target output dimension.

% \begin{figure}
%  \centering
%   \includegraphics[width=\linewidth]{Cros.pdf} 
%     \caption{Figure 4 \textbf{Cross-Period Pattern Interaction.} We devise a causal-enhanced transformer to capture the correlation across different period patterns ($Period_1, \dots, Period_k$). Specifically, we define and apply a strict causal upper triangular matrix (Directed Acyclic Graph) to enforce the acyclic causal constraint of distinct period patterns.}
%   \label{fig4:causal}
% \end{figure}

\begin{figure}
 \centering
  \includegraphics[width=1\linewidth]{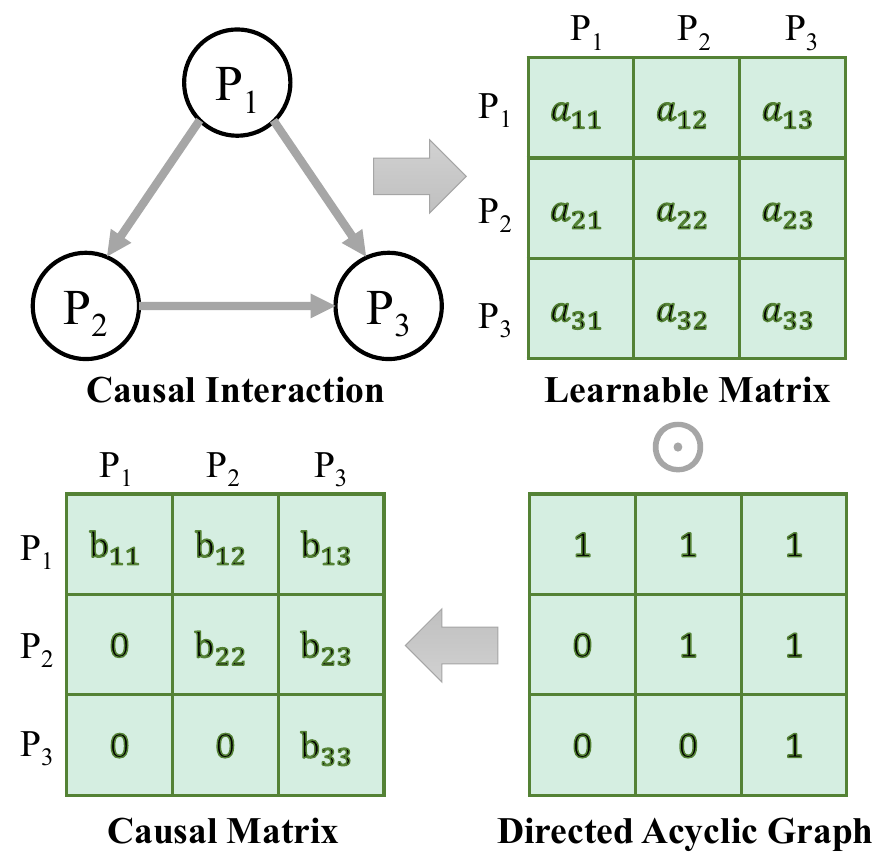} 
    \caption{Figure 4 \textbf{Causal Interaction and Causal Matrix.} We observe that strong period patterns (e.g., $P_1$) exert a dominant causal influence on weak period patterns (e.g., $P_2$,$P_3$), and this interaction is unidirectional.
    Therefore, we define and apply a strict upper triangular matrix (Directed Acyclic Graph) to enforce the acyclic causal constraint of distinct period patterns.}
  \label{fig4:causal}
\end{figure}

\paragraph{Complexity Analysis}

The complexity of the multi-period spatial modeling block is analyzed as follows. The bottleneck projection compresses the feature dimension by 2D convolution, reducing the subsequent computation from full $D$-dimensional operations to quarter-scale $D' = D/4$ ones. The sparse graph convolution performs message passing over only $K$ neighbors per node instead of the full $N$-node graph, reducing the per-layer cost from $O(N_b N T D')$ for dense aggregation to $O(N_b K T D' + N_b T D'^2)$, where $K \ll N$. The global memory bank updates all $N$ nodes once per batch via exponential moving average at cost $O(ND')$; this overhead is negligible during inference since the bank serves as a static lookup table. The output projection maps features back to the target dimension at quarter-scale cost. In total, the block achieves $O(N_b T D^2 + N_b K T D+ND')$ complexity, reducing the quadratic node cost $O(N^2 T D)$ of dense graph convolution to linear and enabling scalable deployment on real-world road networks with thousands of nodes.

\subsection{Cross-Period Pattern Interaction} \label{secte}
We design multi-period temporal modeling and multi-period spatial modeling modules to capture spatio-temporal dependencies for each period pattern.
However, there exist complex causal interactions among various period patterns. We observe an inherent, fixed, and unidirectional causality across diverse period patterns: stronger patterns (e.g., $P_1$) exert a dominant causal influence on weaker ones (e.g., $P_2, P_3$).

To model this causal relationship, we develop a cross-period pattern interaction module. We first combine the spatio-temporal dependencies of all $k$ period patterns into $X_c = (X_1^s, X_2^s, \ldots, X_k^s) \in \mathbb{R}^{k \times N \times T \times D}$, strictly ordered from the strongest pattern $P_1$ to the weakest ones. Breaking this causal ordering leads to a significant drop in model performance, as verified in Section~\ref{crossabl}. As illustrated in Figure 4, we devise a causal mask operation based on a directed acyclic graph (DAG) to enforce information flow only from stronger patterns to weaker ones. This ensures that the representation of each period pattern is derived from itself and stronger period patterns, thereby effectively capturing cross-period pattern interactions.
Then, we generate query $Q$, key $K$, and value $V$ tensors through three separate linear projection layers:
\begin{equation}
Q = W_q(X_c), \quad K = W_k(X_c), \quad V = W_v(X_c)
\end{equation}

Here $W_q$, $W_k$, and $W_v$ denote learnable linear projection matrices with the input and output dimension of D.

Subsequently, we calculate the cross-period similarity matrix between query and key:
\begin{equation}
\text{Sim} = \frac{Q \cdot K^\top}{\sqrt{D}}
\end{equation}

where $\text{Sim} \in \mathbb{R}^{k \times k}$ denotes the similarity scores among different period patterns, and $\sqrt{D}$ is the scaling factor.

Next, we construct the causal matrix $\alpha_{\text{causal}}$ by imposing a scale-constrained mask on learnable parameters: 
\begin{equation}
\begin{split}
\alpha_{\text{causal}} &= \sigma(\alpha) \odot \mathcal{M}_{\text{dag}} \\
% \alpha_{\text{norm}} &= \frac{\alpha_{\text{causal}}}{\sum_{j=1}^K \alpha_{\text{causal},i,j} }
\end{split}
\end{equation}

where $\boldsymbol{\alpha} \in \mathbb{R}^{k \times k}$ is a learnable matrix that adapts cross-scale interaction strengths, $\sigma(\cdot)$ is the sigmoid activation. 
$\mathcal{M}_{\text{dag}} \in \{0, 1\}^{k \times k}$ is an upper triangular mask where $\mathcal{M}_{\text{dag}}[i, j] = 1$ if $i \leq j$, where smaller indices correspond to stronger period patterns, enforcing that higher-energy patterns only attend to lower-energy ones.
The resulting $\alpha_{\text{causal}}$ is a soft causal strength matrix in $[0, 1]$, which preserves acyclicity while adaptively learning the magnitude of causal influence between valid period pattern pairs.

We then fuse the normalized causal weights with the softmax-normalized similarity matrix to obtain the final causal attention weights:
\begin{equation}
\mathcal{W} = \text{Softmax}(\text{Sim}) \odot \alpha_{\text{causal}}
\end{equation}

Finally, we perform weighted aggregation on $V$ using the causal attention weights and sum the results across all period patterns to obtain the final cross-period feature representation $X_{\text{o}} \in \mathbb{R}^{ T \times N \times D}$:
\begin{equation}
\begin{split}
X_{\text{agg}} &= \mathcal{W} \cdot V \\
X_{\text{o}} &= \sum_{k=1}^K X_{\text{agg},k}
\end{split}
\end{equation}

\subsection{Training Strategy}
MP3 is a universal pre-training plugin to enhance the performance of downstream spatio-temporal prediction. It contains a two-step
training strategy: a pre-training step and a downstream spatio-temporal forecasting. In the pre-training step, the goal is to learn multi-period patterns from long-term time series. Downstream spatio-temporal forecasting aims to seamlessly integrate MP3 into existing STGNN backbones, thereby improving their predictive performance and serving as a plug-and-play component.

In the pre-training step, MP3 takes historical long-term time series $X_{\text{long}}$ as input, adopts a regression task for prediction to obtain the long-term period pattern features $\mathbf{Z}$, and employs MAE as the loss function.
\begin{equation}
    \begin{split}
        \boldsymbol{Z} &= f(X_{\text{long}}),\\
    \end{split}
\end{equation}

\begin{figure}[!htb]
 \centering
  \includegraphics[width=\linewidth]{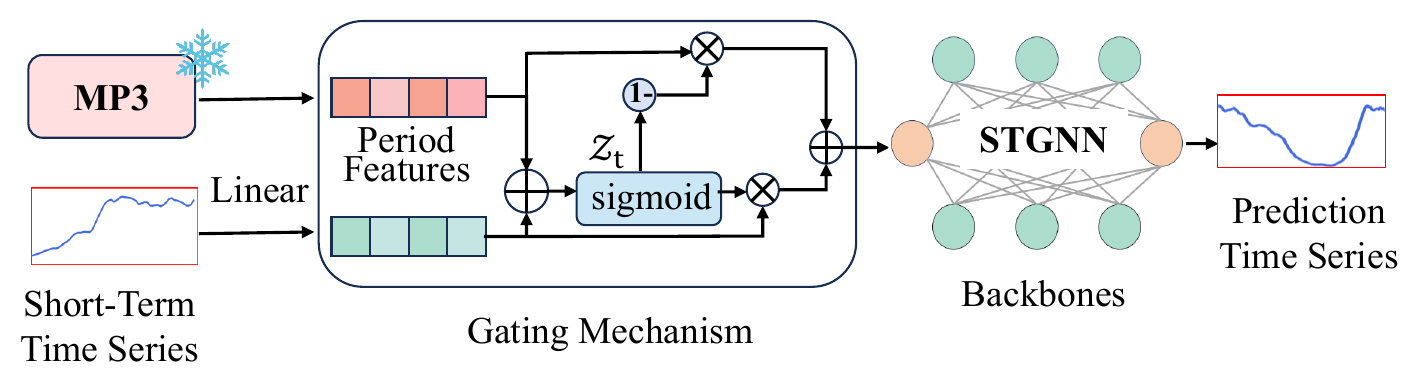} 
    \caption{Figure 5 \textbf{Downstream Spatio-Temporal Forecasting.} As a plug-and-play component, MP3 is frozen and seamlessly integrated with short-term time series features via a gating mechanism to enhance existing STGNN backbones.}
  \label{fig5:fusion}
\end{figure}
 In the downstream spatio-temporal forecasting step, MP3 is frozen and set to eval mode, as shown in Figure 5.
 To integrate the learned multi-period pattern features with short-term observations, a gating mechanism is applied to fuse
 $\mathbf{Z}$ with the short-term raw data $X_{\text{short}} = \left( \mathbf{X}_{t-T'+1}, \mathbf{X}_{t-T'+2}, \dots, \mathbf{X}_t \right)$, where $T'$ is the predefined length of short-term time series. The fusion process is formally defined as:
\begin{equation}
\begin{split}
    \boldsymbol{E}& = \mathrm{Linear}\big(\boldsymbol{X}_\mathrm{short}\big),\\
    \quad \gamma &= \delta\left(\boldsymbol{Z}\mathbf{W}_{h,1} + \mathbf{E}\mathbf{W}_{h,2} + \mathbf{b}_h\right),\\
\mathbf{H} &= \gamma \cdot \boldsymbol{Z} + (1 - \gamma) \cdot \mathbf{E},
\end{split}
\end{equation}

Where $\mathrm{Linear}(\cdot)$ denotes data linear representation, and $\delta$ is a sigmoid activation function. $\mathbf{W}_{h,1}, \mathbf{W}_{h,2} \in \mathbb{R}^{d \times d}$ and $\mathbf{b}_h \in \mathbb{R}^d$ are learnable parameters, enabling the model to autonomously select the period pattern representation learned by MP3. The plugin MP3 effectively mitigates the limitation that downstream models are unable to capture long-term multi-period dependencies.

\section{EXPERIMENT}

\subsection{Experimental Setup}
\subsubsection{DataSet}
We evaluated MP3 on the four most widely used spatio-temporal prediction benchmarks. PEMS03, PEMS04, PEMS07, PEMS08 datasets \cite{song2020spatial} and a large-scale dataset CA, which collect from the California Performance Measurement System (PeMS), all sampled at 5-minute intervals. CA leverages a full traffic network including 9,638 traffic sensors deployed across all nine districts \cite{BigST2024}. 
More detailed description of these five benchmarks can be presented in Table \ref{tab:dataset}. Following standard preprocessing, we applied Z-score normalization to transform raw observations to zero-mean, unit-variance distributions. 
To maintain a consistent comparison with existing works, we adhere to the conventional dataset partitions widely adopted in prior work. 
In the case of the five datasets, the split of training, validation, and testing datasets is 6:2:2. 
% In this work, we ensured the ethical use of these six publicly vailable datasets.

\begin{table}[!h]
    \centering
    \begin{tabular}{llll}
        \hline
        Dataset & Sensors & Timesteps & Time Range \\
        \hline
        PEMS03      & 358   & 26,208 &09/2018-11/2018\\
        PEMS04    & 307   & 16,992 &01/2018-02/2018\\
        PEMS07    & 883   & 28,224 &05/2017-08/2017\\
        PEMS08      & 170   & 17,856 &07/2016-08/2016\\
        CA      & 9,638   & 28,224 &04/2022-06/2022\\
        \hline
    \end{tabular}
\caption{Datasets statistics}  \label{tab:dataset}
\end{table}

 \subsubsection{Implementation Settings}
Our plugin MP3, all baselines, and benchmarks are implemented with PyTorch, run on a computational server with the Ubuntu 22.04.3 operating system, and the server was outfitted with NVIDIA GeForce RTX 3090 GPUs.
For the spatio-temporal forecasting step, each baseline model’s plugin and non-plugin variants use the same learning rate and training epochs, while the backbone parameter settings follow the original settings.
% \begin{table}[htbp]
%   \centering
%   \begin{tabular}{lll}
%     \hline
%     Baseline & Epochs & Learning rate \\
%     \hline
%     GWNet    & 200    & 0.002         \\
%     STGCN    & 200    & 0.001         \\
%     ST\_WA   & 200    & 0.001         \\
%     MSDR     & 200    & 0.0015        \\
%     ST\_Norm & 150    & 0.002         \\
%     \hline
%   \end{tabular}
%   \caption{Baseline Training Hyperparameters}
%   \label{tab:hyperparams}
% \end{table}
For MP3, the observation window $T$ is set to 1248, the prediction horizon $Q$ is set to 12, and the hidden dimension is set to 32. Furthermore, we set parameter k to 3 (selecting the top-3 strongest period patterns by amplitude). We trained the model via the Adam optimizer, configuring a learning rate of 0.001, setting gradient clipping to a maximum value of 5, and running the training for 100 epochs. 
 \subsubsection{Evaluation Metric}
 The evaluation is conducted using three widely-adopted metrics in spatio-temporal prediction \cite{liang2023mixed}: Mean Absolute Error (MAE), Root Mean Square Error (RMSE), and Mean Absolute Percentage Error (MAPE). These metrics are formulated as: 

%Mean Absolute Error (MAE)
\begin{equation}
MAE=\frac{1}{n}\sum_{i=1}^{n}\left|\hat{y_i}-y_i \right|
\end{equation}
%Root Mean Square Error (RMSE)
\begin{equation}
RMSE=\sqrt{\frac{1}{n}\sum_{i=1}^{n}\left(\hat{y_i}-y_i\right)^2} 
\end{equation}
%Mean Absolute Percentage Error (MAPE)
\begin{equation}
MAPE=\frac{100\%}{n}\sum_{i=1}^{n}\left|\frac{\hat{y_i}-y_i}{y_i}\right|
\end{equation}

Here $n$ is the number of test samples, $\hat{y_i}$ and $y_i$ mean the predicted value and the ground truth, respectively.
 \subsubsection{Baselines}
 The MP3 serves as a multi-period pattern learning plugin that can be integrated into various existing spatio-temporal prediction backbones. To validate the enhancement of MP3, we employ five typical spatio-temporal prediction models as backbones.
 \begin{itemize}
     \item \textbf{STGCN \cite{yu2018spatio}:} STGCN is a typical traffic flow prediction method, which designs a sandwich framework consisting of two temporal convolution networks and one graph convolution network, which only leverages the convolution operation to model complex spatio-temporal dependencies. 
     \item \textbf{GWNet \cite{GraphwaveNet_2019}:} GWNet combines dilated causal convolutions to capture long-range temporal dependencies with a novel self-adaptive graph structure to learn the intrinsic and dynamic spatial relationships between nodes in a graph without requiring a predefined structure. 
     \item \textbf{STWA \cite{ST-WA2022}:} STWA maps time series from distinct traffic sensors into stochastic variables, and dynamically designs location and time customized parameters to model complex, evolving traffic patterns.
     \item \textbf{MSDR \cite{MSDR2022}:} MSDR proposes a novel variant of recurrent neural networks to tackle the challenge of capturing long-range spatial dependencies, explicitly uses the hidden states from various historical time intervals as inputs. It also develops a Graph-based MSDR framework that integrates MSDR with graph neural networks.
     \item \textbf{STNorm \cite{deng2021st}:} STNorm introduces two modular components: Temporal Normalization and Spatial Normalization. TN separates time series into high- and low-frequency components to enhance the former, while the SN module isolates local, variable-specific components from the raw data.
 \end{itemize}
 
 For a fair comparison, we benchmark MP3 against GPT-ST \cite{li2023gptst}, a competitive spatio-temporal modeling plugin that utilizes a specialized parameter learning strategy for spatio-temporal data. To ensure a fair evaluation, both the backbones and MP3-enhanced variants are configured using the optimal hyperparameter settings from the backbones' official implementations.

{\small
\begin{table*}[htbp]
  \centering
  \caption{Experimental results of baseline models and their enhanced versions with GPT-ST and MP3 on four PEMS datasets. Best results in each group are highlighted in bold.}
  \label{tab:results}
  \begin{tabular}{lcccccccccccc}
    \toprule
    \multirow{2}{*}{Model} & \multicolumn{3}{c}{PEMS03} & \multicolumn{3}{c}{PEMS04} & \multicolumn{3}{c}{PEMS07} & \multicolumn{3}{c}{PEMS08} \\
    \cmidrule(lr){2-4} \cmidrule(lr){5-7} \cmidrule(lr){8-10} \cmidrule(lr){11-13}
     & MAE & RMSE & MAPE & MAE & RMSE & MAPE & MAE & RMSE & MAPE & MAE & RMSE & MAPE \\
    \midrule

 % STGCN Group
    STGCN     & 17.49 & 30.12 & 17.15\% & 19.88 & 31.19 & 13.63\% & 22.79 & 36.18 & 9.93\% & 16.25 & 25.13 & 11.07\% \\ 
    \quad+GPT-ST & 16.83 & 28.48 & 18.13\% & 20.08 & 35.46 & 13.22\% & 22.56 & 36.91 & 9.84\% & 16.24 & 25.93 & 10.45\% \\ 
    \quad+MP3 & \textbf{15.99} & \textbf{26.02} & \textbf{15.14\%} & \textbf{19.03} & \textbf{29.61} & \textbf{13.20\%} & \textbf{20.20} & \textbf{32.86} & \textbf{8.59\%} & \textbf{14.83} & \textbf{22.81} & \textbf{9.87\%} \\ 
    \quad\quad$\Delta$ & \cellcolor{green!20}-1.50 & \cellcolor{green!20}-4.10 & \cellcolor{green!20}-2.01\% & \cellcolor{green!20}-0.85 & \cellcolor{green!20}-1.58 & \cellcolor{green!20}-0.43\% & \cellcolor{green!20}-2.59 & \cellcolor{green!20}-3.32 & \cellcolor{green!20}-1.34\% & \cellcolor{green!20}-1.42 & \cellcolor{green!20}-2.32 & \cellcolor{green!20}-1.20\% \\
    \midrule
    
    % GWNet Group
    GWNet & 15.27 & 23.98 & 14.98\% & 18.74 & 28.64 & 12.87\% & 20.35 & 31.98 & 8.60\% & 14.40 & 23.39 & 9.21\% \\ 
    \quad+GPT-ST & 14.90 & 26.92 & 15.16\% & 19.01 & 32.95 & 12.81\% & 19.93 & 33.20 & 8.33\% & 13.96 & 22.83 & 9.16\% \\ 
    \quad+MP3 & \textbf{14.89} & \textbf{23.61} & \textbf{14.44\%} & \textbf{18.48} & \textbf{28.33} & \textbf{12.80\%} & \textbf{18.89} & \textbf{30.54} & \textbf{8.09\%} & \textbf{13.77} & \textbf{21.43} & \textbf{9.14\%} \\ 
    \quad\quad$\Delta$ & \cellcolor{green!20}-0.38 & \cellcolor{green!20}-0.37 & \cellcolor{green!20}-0.54\% & \cellcolor{green!20}-0.26 & \cellcolor{green!20}-0.31 & \cellcolor{green!20}-0.07\% & \cellcolor{green!20}-1.46 & \cellcolor{green!20}-1.44 & \cellcolor{green!20}-0.51\% & \cellcolor{green!20}-0.63 & \cellcolor{green!20}-1.96 & \cellcolor{green!20}-0.07\% \\
    \midrule

    % STWA Group
    STWA     & 16.14 & 25.68 & 15.65\% & 19.92 & 30.02 & 13.82\% & 21.41 & 34.20 & 9.23\% & 16.16 & 25.80 & 10.55\% \\ 
    \quad+GPT-ST & \textbf{15.46} & 27.72 & 15.31\% & 19.50 & 33.26 & 13.08\% & 20.62 & 34.52 & 8.84\% & 15.29 & 24.50 & \textbf{9.81\%} \\ 
    \quad+MP3  & 15.88 & \textbf{25.13} & \textbf{15.13\%} & \textbf{19.17} & \textbf{29.11} & \textbf{12.90\%} & \textbf{20.23} & \textbf{32.07} & \textbf{8.82\%} & \textbf{15.03} & \textbf{22.71} & 10.19\% \\ 
    \quad\quad$\Delta$ & \cellcolor{green!20}-0.26 & \cellcolor{green!20}-0.55 & \cellcolor{green!20}-0.52\% & \cellcolor{green!20}-0.75 & \cellcolor{green!20}-0.91 & \cellcolor{green!20}-0.92\% & \cellcolor{green!20}-1.18 & \cellcolor{green!20}-2.13 & \cellcolor{green!20}-0.41\% & \cellcolor{green!20}-1.13 & \cellcolor{green!20}-3.07 & \cellcolor{green!20}-0.36\% \\
    \midrule

    % MSDR Group
    MSDR      & 15.76 & 24.80 & 15.51\% & 20.39 & 30.93 & 14.22\% & 22.18 & 34.28 & 9.82\% & 16.33 & 24.69 & 10.60\% \\ 
    \quad+GPT-ST & 15.87 & 26.60 & 16.02\% & 19.86 & 33.58 & \textbf{13.50\%} & 21.36 & 34.54 & 9.32\% & 15.92 & 25.03 & 10.33\% \\ 
    \quad+MP3  & \textbf{15.53} & \textbf{24.75} & \textbf{15.09\%} & \textbf{19.64} & \textbf{30.07} & 13.85\% & \textbf{20.99} & \textbf{32.90} & \textbf{9.26\%} & \textbf{15.18} & \textbf{23.32} & \textbf{10.28\%} \\ 
    \quad\quad $\Delta$ & \cellcolor{green!20}-0.23 & \cellcolor{green!20}-0.05 & \cellcolor{green!20}-0.42\% & \cellcolor{green!20}-0.75 & \cellcolor{green!20}-0.86 & \cellcolor{green!20}-0.37\% & \cellcolor{green!20}-1.19 & \cellcolor{green!20}-1.38 & \cellcolor{green!20}-0.56\% & \cellcolor{green!20}-1.15 & \cellcolor{green!20}-1.37 & \cellcolor{green!20}-0.32\% \\
    \midrule

    % STNorm Group
    STNorm     & 15.69 & 24.88 & 15.10\% & 19.54 & 29.81 & 13.51\% & 21.06 & 33.46 & 9.08\% & 15.82 & 23.74 & 10.05\% \\ 
    \quad+GPT-ST & \textbf{15.50} & 27.52 & 15.85\% & 19.15 & 33.55 & \textbf{12.69\%} & 20.25 & 34.60 & 8.54\% & 15.10 & 25.10 & 10.03\% \\ 
    \quad+MP3  & 15.59 & \textbf{24.86} & \textbf{14.66\%} & \textbf{18.85} & \textbf{28.74} & 13.07\% & \textbf{20.11} & \textbf{32.12} & \textbf{8.54\%} & \textbf{14.73} & \textbf{22.40} & \textbf{9.85\%} \\ 
    \quad\quad $\Delta$ & \cellcolor{green!20}-0.10 & \cellcolor{green!20}-0.02 & \cellcolor{green!20}-0.44\% & \cellcolor{green!20}-0.69 & \cellcolor{green!20}-1.07 & \cellcolor{green!20}-0.44\% & \cellcolor{green!20}-0.95 & \cellcolor{green!20}-1.34 & \cellcolor{green!20}-0.54\% & \cellcolor{green!20}-1.09 & \cellcolor{green!20}-1.34 & \cellcolor{green!20}-0.20\% \\
    
    \bottomrule
  \end{tabular}
  \label{Experiment}
\end{table*}
}

\begin{figure*}[!htb]
 \centering
\includegraphics[width=1\linewidth]{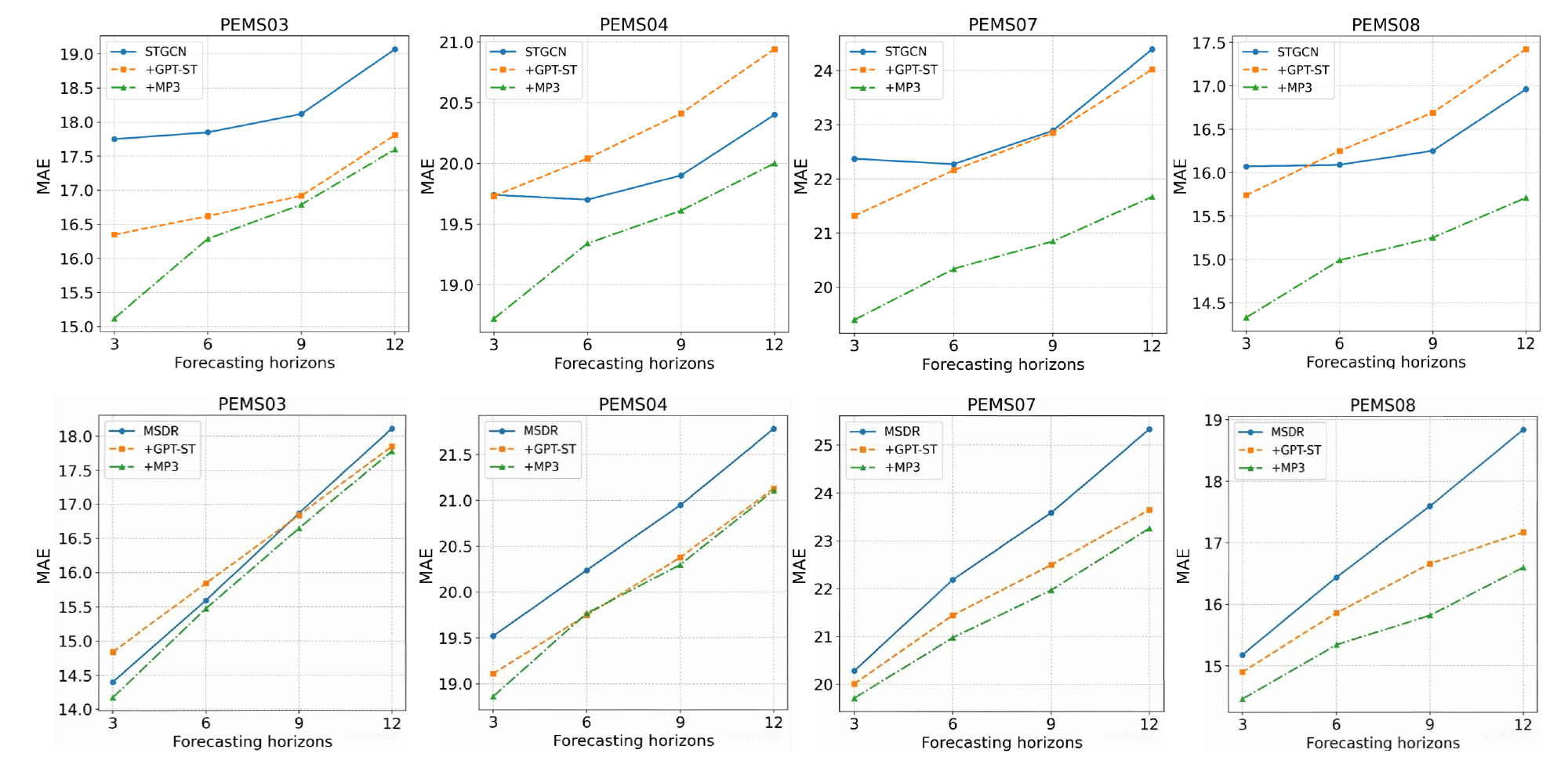} 
    \caption{Figure 6 \textbf{ Performance Comparison of Forecasting Horizons across Four Datasets on STGCN and MSDR Backbones}}
  \label{fig6:temporalAB}
\end{figure*}
 
 \subsection{Performance Analysis}
 \subsubsection{Overall Performance}
Table \ref{Experiment} provides the experimental outcomes of MP3 evaluated over five representative backbones and four spatio-temporal datasets. The results consistently indicate that the integration of MP3 yields substantial performance improvements across all evaluated backbones. Notably, in the majority of cases, backbones enhanced with MP3 achieve state-of-the-art results, surpassing not only their base models but also those augmented with the competing GPT-ST framework. A comprehensive analysis of MP3's performance enhancements is conducted from key dimensions:

\begin{itemize}
    \item \textbf{Consistent Performance Enhancement with MP3.} The experimental results demonstrate that MP3 yields consistent performance gains across diverse baseline methods, and such a positive effect is not confined to any specific type of baseline, which manifests the generalization ability of MP3. We attribute this improvement to the multi-period knowledge distillation and multi-period pattern learning of long-term time series in the pre-training stage.
    \item \textbf{Comparative Analysis of Different Backbones.} On the large-scale PEMS07 dataset, MP3 delivers consistent performance gains across all backbone networks and all three evaluation metrics. Moreover, relative to recent methods such as STNorm and MSDR, MP3 yields notably greater performance gains when integrated with classical baselines like STGCN. Specifically, integrating MP3 with STNorm yields gains of 0.95 in MAE, 1.34 in RMSE, and 0.54\% in MAPE, whereas integrating MP3 with STGCN leads to larger gains of 2.59 in MAE, 3.32 in RMSE, and 1.34\% in MAPE on PEMS07 dataset. This finding can be attributed to the fact that advanced baselines such as STNorm are already carefully designed to capture diverse spatio-temporal patterns, thereby encoding rich prior knowledge and leaving comparatively limited room for additional signals from pre-training. In contrast, classical methods such as STGCN, due to their simpler design and more restricted representational capacity, can exploit the auxiliary information provided by MP3 more effectively, leading to substantially improved modeling of complex dependencies.
    \item \textbf{Comparison with Pre-training Methods.} Except for evaluating MP3 against STGNNs trained from scratch, we also benchmark MP3 against GPT-ST, a pre-training method designed to learn spatio-temporal heterogeneity. Different from GPT-ST, our MP3 mainly focuses on the multi-period feature and multi-period pattern learning, which achieves superior performance to GPT-ST on all datasets and three evaluation metrics. Furthermore, MP3 validates an even more notable performance advantage on the large-scale dataset PEMS07. This highlights the broader scalability of MP3.
\end{itemize}

% \begin{figure} [!htb]
%  \centering
% \includegraphics[width=1\linewidth]{cost.pdf} 
%     \caption{Figure 6 \textbf{ Comparison in Terms of MAE, Training Time, and Parameter Size of Backbones and Their +MP3 Variants.}}
%   \label{fig7:cost}
% \end{figure}

\subsubsection{Performance on Different Forecasting Horizons}
Figure 6 reports the MAE results of forecasting horizons 3, 6, 9, 12 cross four datasets on STGCN and MSDR backbones. The key findings are threefold. First, MP3 consistently enhances the performance of STGCN and MSDR across all horizons and surpasses GPT-ST, further validating the robustness of the results in Table \ref{Experiment}. Second, the improvement of MP3 on MSDR is less pronounced than on STGCN, as MSDR already accounts for complex long-term spatio-temporal dependencies, which is consistent with Table \ref{Experiment}. Third, MP3 has a greater impact on short-horizon predictions, achieving remarkable gains, particularly at Horizon 3. The longer window of Horizon 12 lets the backbone capture more period patterns by itself, while the shorter window at Horizon 3 limits such ability, making the period context(e.g., phase in long-period patterns ) from MP3 more impactful. 
% \subsubsection{Computational Cost}
% As shown in Figure 6, the integration of MP3 consistently reduces the MAE across all backbone models. Notably, the “+MP3” variants achieve improved prediction accuracy with only marginal changes in parameter size and negligible additional training time. This indicates that MP3 provides an effective and lightweight enhancement, yielding a favorable trade-off between efficiency and effectiveness.
% \begin{figure*} [htbp]
%  \centering
% \includegraphics[width=1\linewidth]{ablation_study.pdf} 
%     \caption{Figure 8 \textbf{Experimental result of Cross-Period Pattern Interaction Ablation.}}
%   \label{fig7:CrossAB}
% \end{figure*}
\subsection{Ablation Study}
We carry out ablation studies with multi-period temporal modeling ablation, cross-period pattern interaction ablation, and other variants to investigate how each module affects the performance of MP3 on the PEMS03 and PEMS08 datasets and STGCN and GWNet backbones.

\subsubsection{Multi-Period Temporal Modeling Ablation} To validate the impact of multi-period temporal modeling, we introduce two variants for comparison:
\begin{itemize}
    \item \textbf{with 2Dc:} Replace the edge convolution with 2D convolution.
    \item \textbf{with Linear:} Substitute the edge convolution with a linear layer.
\end{itemize}

Figure 7 presents the experimental results of multi-period temporal modeling variants on PEMS03 and PEMS08 using STGCN and GWNet as backbones. The results indicate that the use of 2‑D convolution substantially degrades the performance of the model, highlighting the superiority of the proposed edge convolution method. Furthermore, employing Linear, a widely adopted and lightweight temporal dependency modeling approach, also results in a noticeable decline in predictive performance.

\begin{figure*}[!htb]
 \centering
\includegraphics[width=1\linewidth]{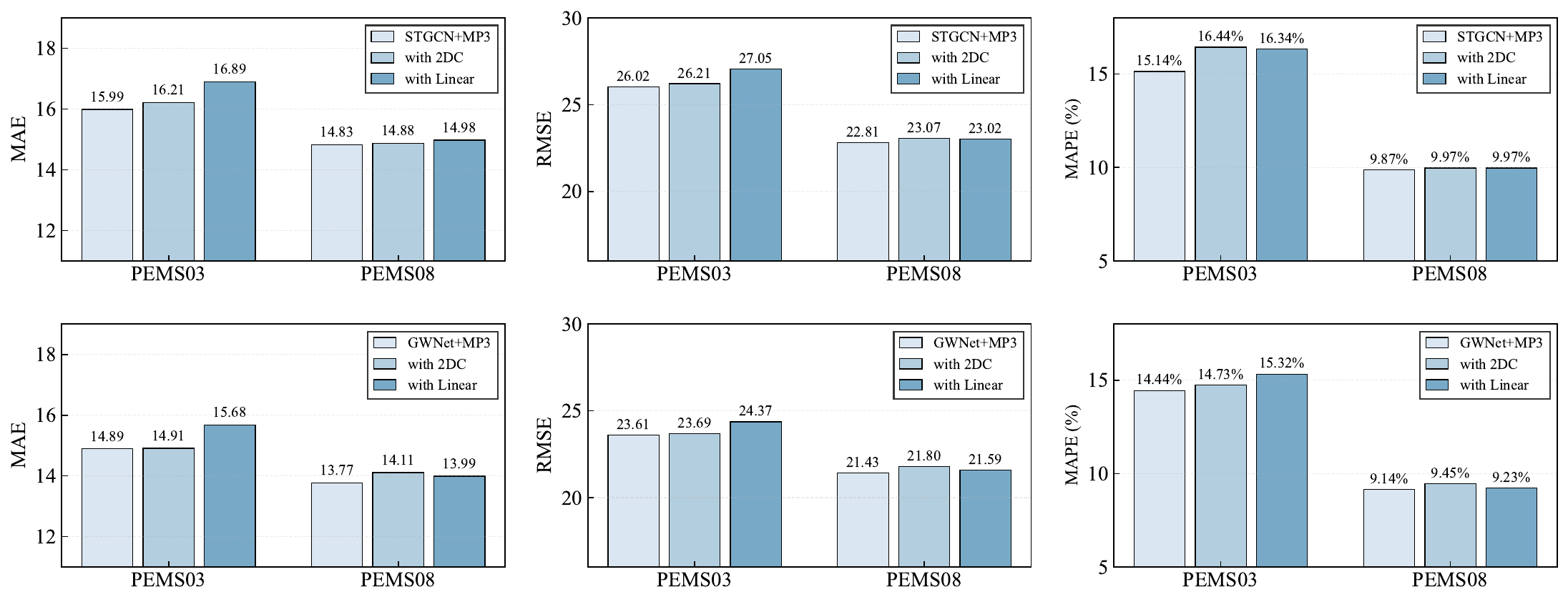} 
    \caption{Figure 7 \textbf{Experimental Results of Multi-Period Temporal Modeling Ablation}}
  \label{fig7:temporalAB}
\end{figure*}

\subsubsection{Cross-Period Pattern Interaction Ablation} \label{crossabl}
To evaluate the effectiveness of cross-period pattern interaction, we design three ablation variants:
\begin{itemize}
    \item \textbf{w/o Cro:} Remove cross-period pattern interaction module.
     \item \textbf{w/o CaM:} Remove the directed acyclic graph mask matrix, and leverage the traditional transformer.
    \item \textbf{w/ Perturb:} Manually reverse the natural strength order of period patterns by swapping the feature tensors of top-1 and top-2 period patterns before the DAG causal mask module, breaking the strong-to-weak unidirectional causal interactions.
\end{itemize}

\begin{table}[htbp]
    \centering
    \caption{Experimental result of cross-period pattern interaction ablation with STGCN and GWNet backbones on PEMS03 and PEMS08 datasets}
    \label{tab:causal_order_ablation}
    \small % 和论文其他表格字体统一
    \setlength{\tabcolsep}{3pt} % 组内基础列间距，兼顾紧凑与可读性
    % 核心修改：两组3列之间插入10pt专属间距，彻底拉开两个大列
    \begin{tabular}{l *{3}{c} @{\hspace{4pt}} *{3}{c}}
        \toprule
        \multirow{2}{*}{Model} & \multicolumn{3}{c}{PEMS03} & \multicolumn{3}{c}{PEMS08} \\
        \cmidrule(lr){2-4} \cmidrule(lr){5-7}
        & MAE & RMSE & MAPE & MAE & RMSE & MAPE \\
        \midrule
        % \textbf{MSDR} \\
        % \quad+MP3 & 15.53 & 24.75 & 15.09\% & 15.18 & 23.32 & 10.28\% \\
        % \quad w/ Perturb & 16.32 & 26.06 & 16.55\% & 15.51 & 23.70 & 10.31\% \\
        % \midrule
        % \textbf{STWA} \\
        % \quad+MP3 & 15.88 & 25.13 & 15.13\% & 15.03 & 22.71 & 10.19\% \\
        % \quad w/ Perturb & 16.88 & 26.49 & 17.10\% & 15.16 & 22.84 & 10.25\% \\
        % \midrule
        \textbf{STGCN} \\
        \quad+MP3 & \textbf{15.99} & \textbf{26.02} & \textbf{15.14}\% & \textbf{14.83} & \textbf{22.81} & \textbf{9.87\%} \\
        \quad w/o Cro    & 16.40 & 26.48 & 16.74\% & 14.97 & 22.99 & 10.30\% \\
        \quad  w/o CaM     & 16.40 & 26.38 & 16.49\% & 14.99 & 23.19 & 9.92\% \\
        \quad w/ Perturb & 16.94 & 26.82 & 17.67\% & 15.01 & 23.16 & 9.97\% \\
        \midrule
        \textbf{GWNet} \\
        \quad +MP3 & \textbf{14.89} & \textbf{23.61} & \textbf{14.44\% }& \textbf{13.77} & \textbf{21.43} & \textbf{9.14\%} \\
        \quad w/o Cro & 15.91 & 24.77 & 15.60\% & 14.00 & 21.76 & 9.38\% \\
        \quad  w/o CaM     & 15.03 & 23.62 & 14.49\% & 13.86 & 21.50 & 9.18\% \\
        \quad w/ Perturb & 15.35 & 23.86 & 15.54\% & 13.91 & 21.63 & 9.15\% \\
        \bottomrule
    \end{tabular}
\end{table}

Table \ref{tab:causal_order_ablation} provides the experimental results of cross-period pattern interaction variants on PEMS03 and PEMS08 using STGCN and GWNet as backbones. The results show that 'w/o Cro' significantly degrades model performance, highlighting the importance of modeling cross‑period pattern interactions. 'w/o CaM' also impairs performance, implying the causal association across different period patterns.
% \subsubsection{Causal Order of Period Patterns Ablation} \label{crossabl}
% To verify the correctness of the core causal hypothesis for our cross-period pattern interaction design, we design one targeted ablation variant:
% \begin{itemize}
%     \item \textbf{w/ Perturb:} Manually reverse the natural strength order of period patterns by swapping the feature tensors of top-1 and top-2 period patterns before the DAG causal mask module, breaking the predefined strong-to-weak unidirectional causal flow.
% \end{itemize}
Moreover,
% Table \ref{tab:causal_order_ablation} provides the experimental results on PEMS03 and PEMS08 across four backbones. 
The results show that 'w/o Perturb' consistently degrades model performance under all settings, which highlights that the inherent strength order of period patterns is the core foundation of cross-period interaction, and provides solid empirical support for our DAG-based causal mask design.

\subsubsection{Other Ablation} To validate the importance of other modules, we consider three ablated versions:

\begin{itemize}
    \item \textbf{w/o Cem:} Remove context convolution for temporal downsampling module.
    \item \textbf{w/o Spa:} Delete multi-period spatial modeling module.
    \item \textbf{with Fix:} In this variant, we replace the dynamic period patterns with fixed period values by removing the multi-period identification and main period selection modules, followed by reshaping the time series with constant periods of 14, 30, and 60.
   
\end{itemize}

\begin{table}[htbp]
    \centering
    \caption{Experimental result of other ablation with STGCN and GWNet backbones on PEMS03 and PEMS08 datasets}
    \label{tab:other_ablation}
    \small % 和论文其他表格字体统一
    \setlength{\tabcolsep}{3pt} % 组内基础列间距，兼顾紧凑与可读性
    % 核心修改：两组3列之间插入10pt专属间距，彻底拉开两个大列
    \begin{tabular}{l *{3}{c} @{\hspace{4pt}} *{3}{c}}
        \toprule
        \multirow{2}{*}{Model} & \multicolumn{3}{c}{PEMS03} & \multicolumn{3}{c}{PEMS08} \\
        \cmidrule(lr){2-4} \cmidrule(lr){5-7}
        & MAE & RMSE & MAPE & MAE & RMSE & MAPE \\
        \midrule
         \textbf{STGCN} \\
       \quad +MP3   & \textbf{15.99} & \textbf{26.02} & \textbf{15.14\% }& \textbf{14.83} & \textbf{22.81} & \textbf{9.87\%} \\
       \quad w/o Cem   & 16.78 & 26.83 & 16.93\% & 15.04 & 23.19 & 10.08\% \\
       \quad w/o Spa       & 16.44 & 26.48 & 16.72\% & 14.92 & 22.99 & 9.94\% \\
       \quad with Fix    & 16.34 & 26.66 & 15.90\% & 14.91 & 22.98 & 9.94\% \\
        \midrule
         \textbf{GWNet} \\
         \quad +MP3 & \textbf{14.89} & \textbf{23.61} & \textbf{14.44\% }& \textbf{13.77} & \textbf{21.43} & \textbf{9.14\%} \\
        \quad w/o Cem   & 15.27 & 23.99 & 15.49\% & 13.94 & 21.55 & 9.50\% \\
        \quad w/o Spa       & 15.21 & 24.13 & 14.75\% & 13.92 & 21.62 & 9.24\% \\
        \quad with Fix    & 15.27 & 23.68 & 14.92\% & 13.97 & 21.77 & 9.24\% \\
        \bottomrule
    \end{tabular}
\end{table}

Table \ref{tab:other_ablation}
present the experimental result of the other ablation study. From Table \ref{tab:other_ablation}, on the PEMS03 and PEMS08 datasets with STGCN and GWNet as backbones, we can gain the following observations:
\begin{itemize}
    \item  "w/o Cem" has a substantial impact on experimental results, leading to a decrease of 0.79 and 0.21 in MAE on PEMS03 and PEMS08 with STGCN as backbone. This indicates that the context convolution for the temporal downsampling module plays a critical role in strengthening the model's overall performance. 
    \item  "w/o Spa" results in comparable performance degradation, with MAE decreasing by 0.45 and 0.32 on PEMS03 with STGCN and GWNet as backbones. This reveals that multi-period spatial pattern learning is important for the performance improvement of the MP3 model.
    % \item "with 2Dc" and "with Lin" suffer from a performance reduction of 0.22 and 0.9 in MAE, respectively, which manifests that the edge convolution can more effectively capture the correlation of inter-period and intra-period variation. 
    \item "with Fix" leads to the performance degradation of 0.35 in MAE on PEMS03 with STGCN as backbone, which indicates that the period feature is dynamic and diverse.
%     \item "w/o Cam" causes MAE to decrease by 0.41, which implies that the causal mask is significant for cross-period pattern interaction and identifies that Period patterns in time also follow
% unidirectional causal dependencies, where past patterns influence subsequent future patterns.
\end{itemize}
 
\begin{figure*}[!htb]
 \centering
\includegraphics[width=1\linewidth]{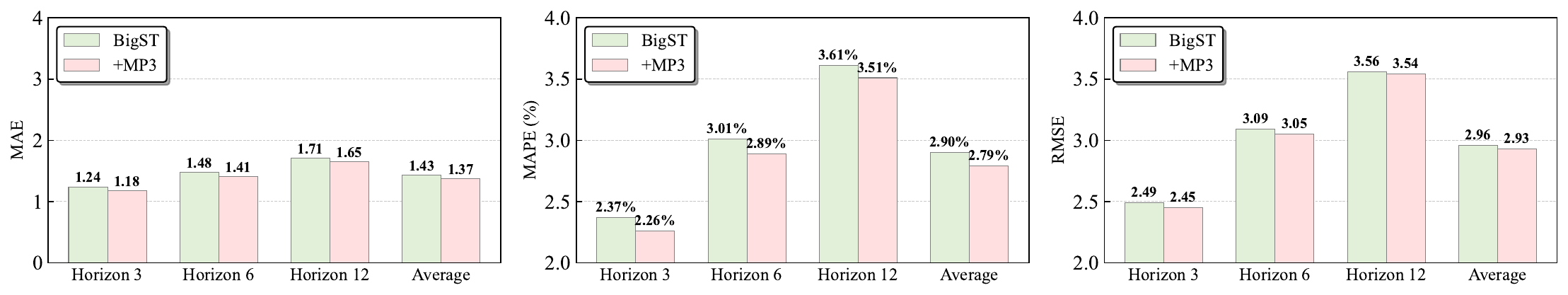} 
    \caption{Figure 8 \textbf{Experimental Results on Large Scale Dataset CA}}
  \label{fig6:BigCA}
\end{figure*}

\subsection{Parameter Sensitivity Analysis}
To evaluate the impact of the input window length \( T \) on model performance and efficiency, we conducted experiments with four different window sizes: 720, 1248, 1800, and 2040. As shown in the Table \ref{parameter_t}, increasing \( T \) leads to moderate improvements in prediction accuracy, with MAE decreasing from 14.18 to 13.71, RMSE from 21.97 to 20.97, and MAPE from 9.34\% to 8.88\%. However, the training time per epoch increases substantially from 229 seconds to 515 seconds, indicating a clear trade-off between predictive performance and computational cost. Considering both aspects, we select \( T = 1248 \) as the optimal setting, as it achieves competitive accuracy while maintaining a moderate training time of 385 seconds per epoch, offering a favorable balance between effectiveness and efficiency.

\begin{table}[htbp] 
\centering
\caption{Performance comparison under different input window lengths}
\begin{tabular}{c c c c c}
\toprule
$T$ & MAE & RMSE & MAPE & Training time (s/epoch) \\
\midrule
720  & 14.18 & 21.97 & 9.34\% & 229  \\
1248 & 13.77 & 21.43 & 9.14\% & 385  \\
1800 & 13.75 & 21.21 & 9.07\% & 430  \\
2040 & 13.71 & 20.97 & 8.88\% & 515  \\
\bottomrule
\end{tabular}\label{parameter_t}
\end{table}

\subsection{Scalability for Large Scale Spatio-Temporal Network}
To validate MP3's superior scalability for large-scale traffic networks, we implement the experiment on the CA dataset and set BigST\cite{BigST2024} as the backbone.
The experimental results are depicted in Figure 8. We compare BigST with the enhanced model by MP3 on Horizon 3 (15 minutes), Horizon 6 (30 minutes), Horizon 12 (1 hour), and the Average of all 12 horizons with the three metrics.
From Figure 8, we draw the following findings: 
\begin{itemize}
    \item MP3 consistently outperforms the baseline across all three evaluation metrics and time horizons, with particularly notable improvements on MAE and MAPE. We attribute this performance gain to the model's ability to distill multi-period knowledge and capture multi-period patterns from long-term traffic time series during pre-training, which are not considered in the BigST model.
    \item MP3 presents strong scalability. It consistently obtains performance improvements not only on small-scale datasets such as PEMS03, PEMS04, PEMS07, and PEMS08 with hundreds of nodes but also on the large-scale CA dataset with thousands of nodes, manifesting its excellent scalability across diverse traffic network scenarios.
\end{itemize}

\begin{figure*}[!htb]
    \centering
    \subfloat[STGCN on PEMS04]{
        \includegraphics[width=0.3\textwidth]{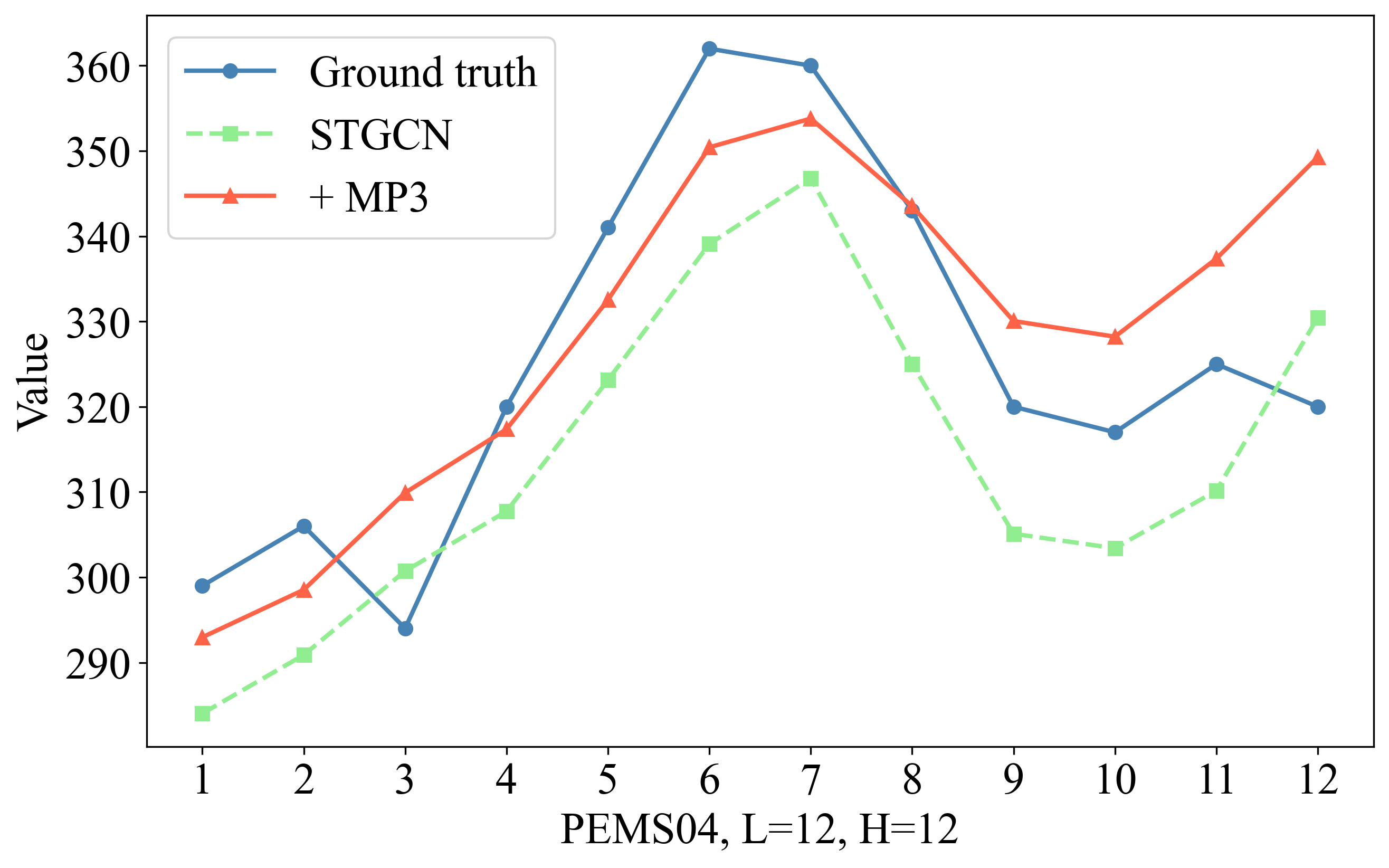}
        \label{fig:sub1}
    }
    \hfill
    \subfloat[STGCN on PEMS08]{
        \includegraphics[width=0.3\textwidth]{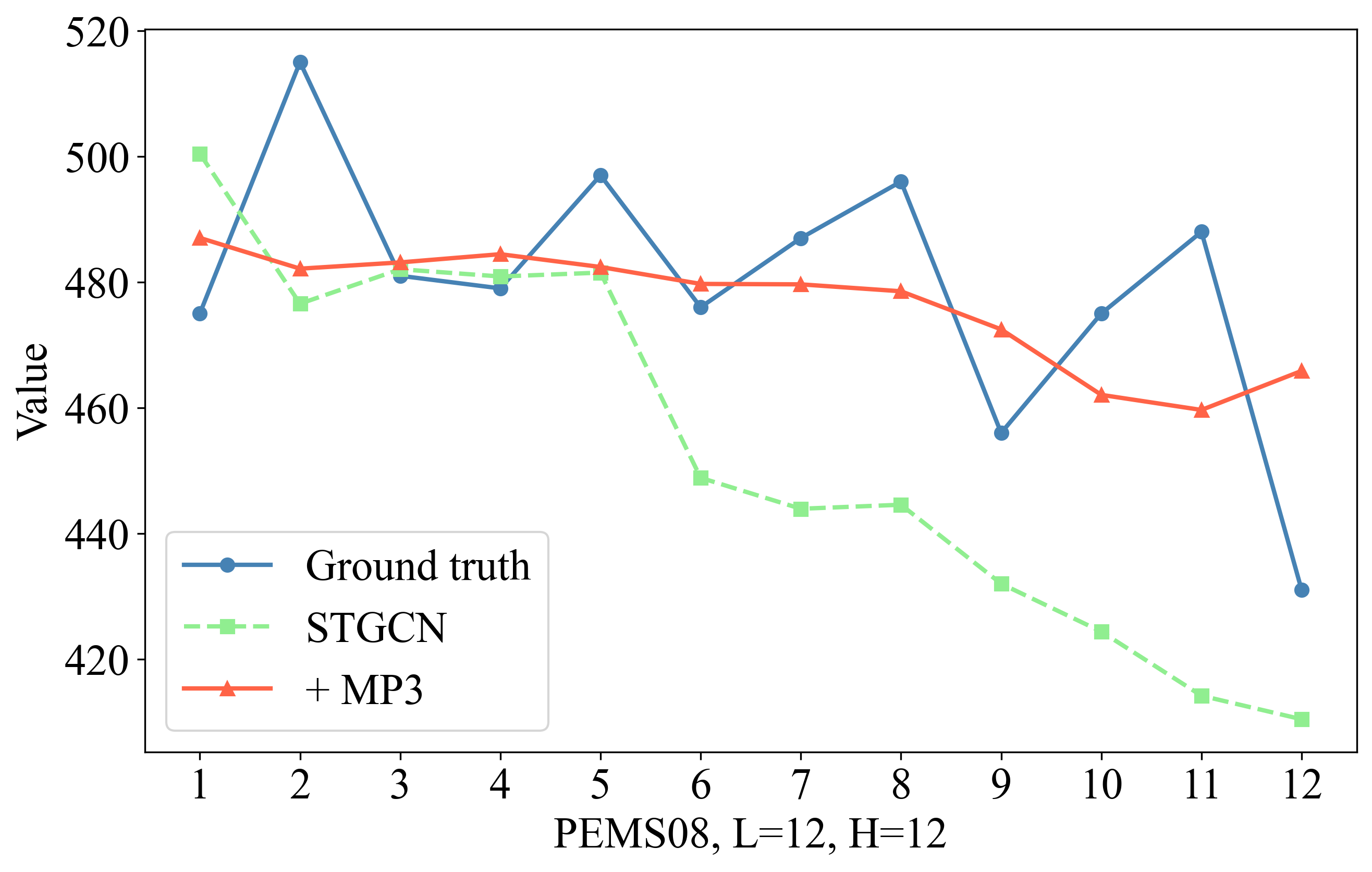}
        \label{fig:sub2}
    }
    \hfill
    \subfloat[MSDR on PEMS04]{
        \includegraphics[width=0.3\textwidth]{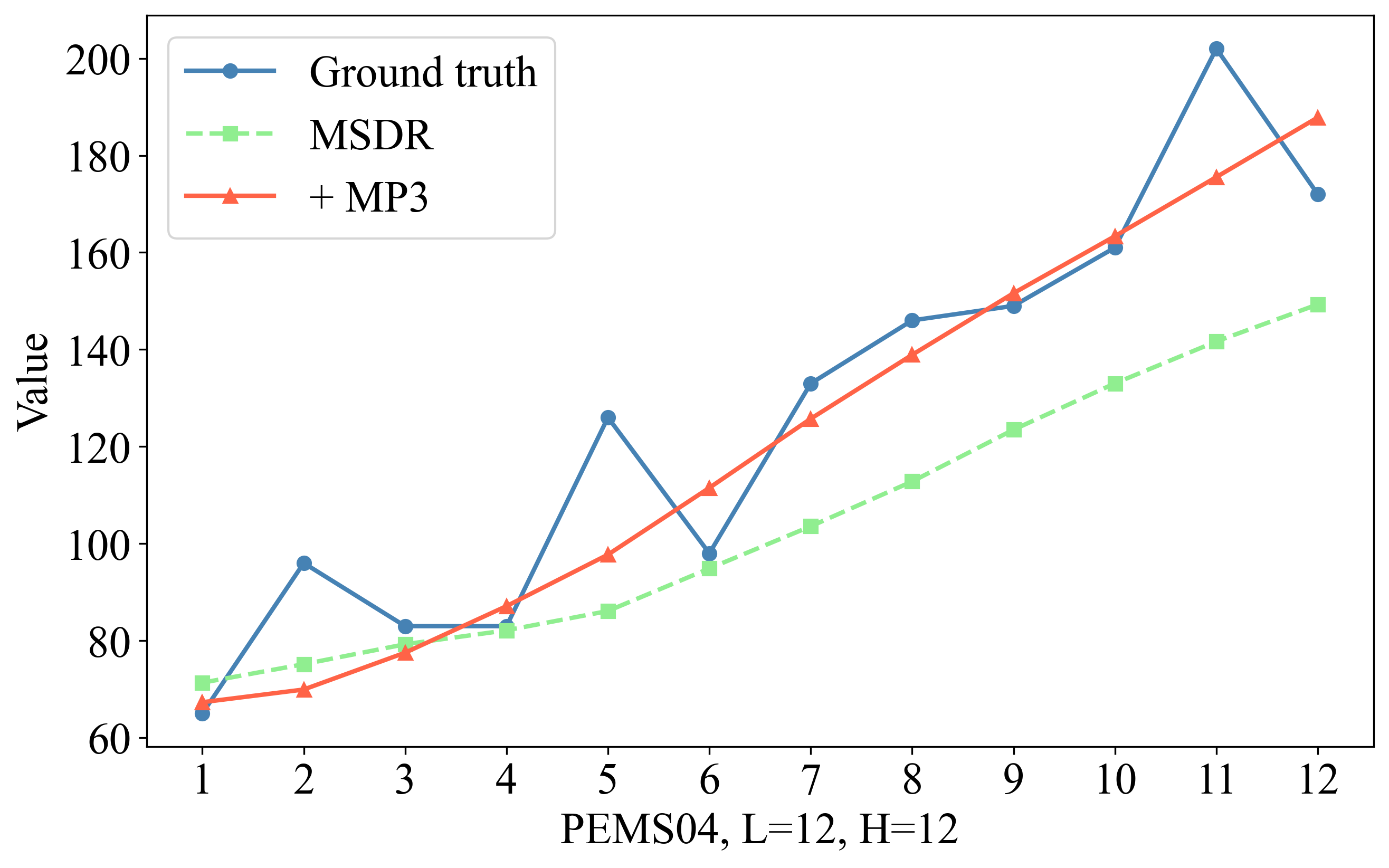}
        \label{fig:sub3}
    }

    % \vspace{\baselineskip}

    \subfloat[MSDR on PEMS08]{
        \includegraphics[width=0.3\textwidth]{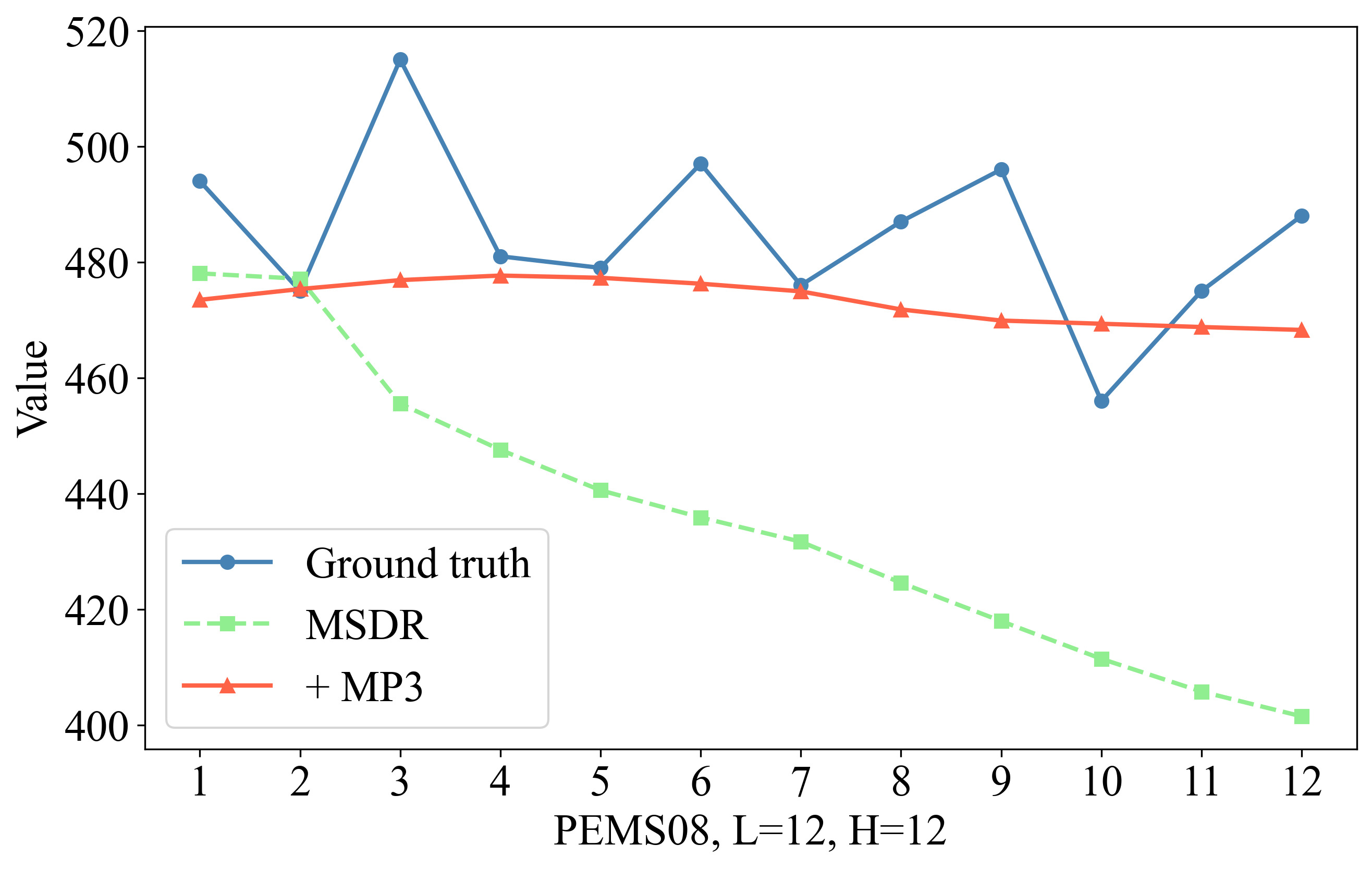}
        \label{fig:sub4}
    }
    \hfill
    \subfloat[STWA on PEMS04]{
        \includegraphics[width=0.3\textwidth]{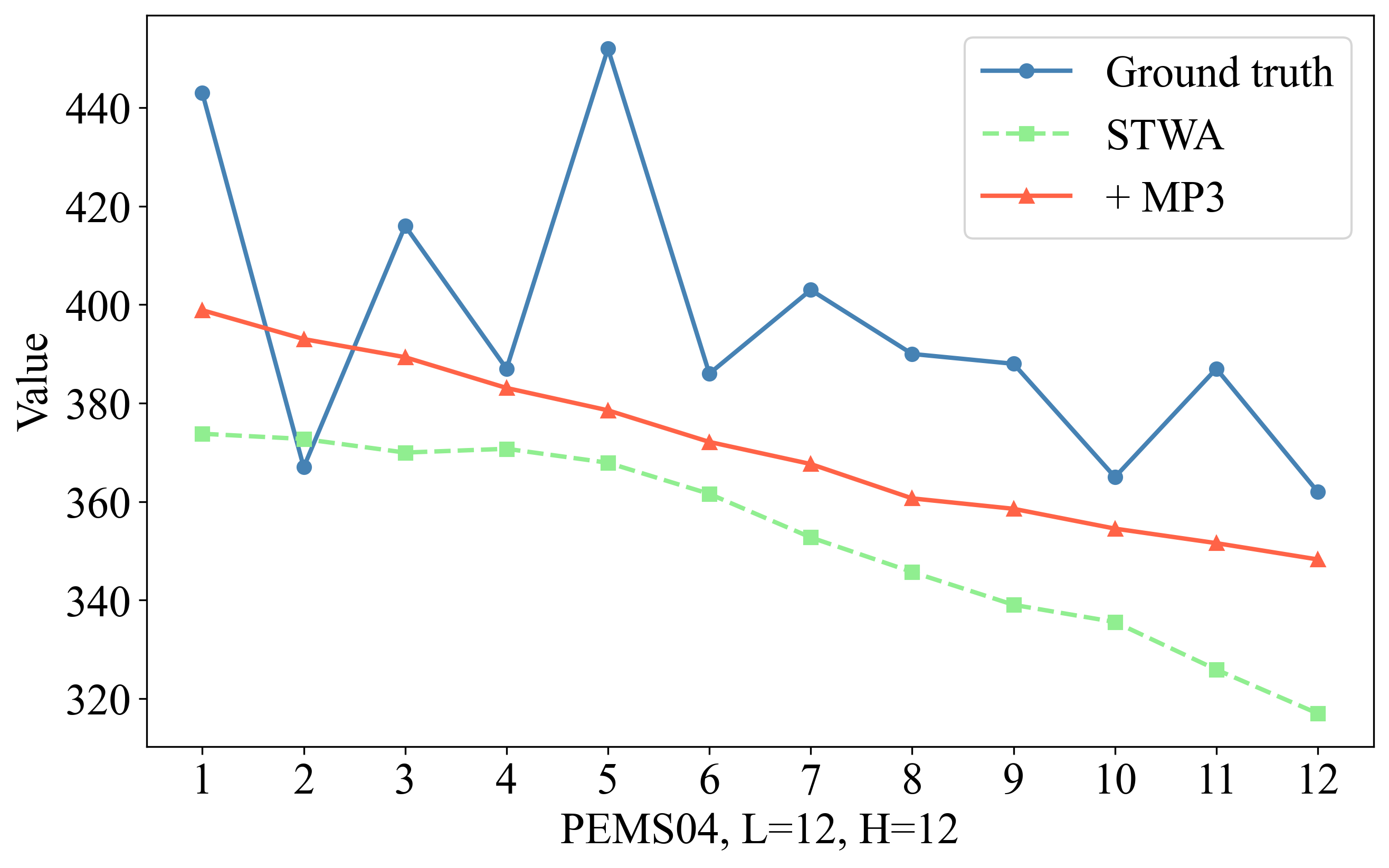}
        \label{fig:sub5}
    }
    \hfill
    \subfloat[STWA on PEMS08]{
        \includegraphics[width=0.3\textwidth]{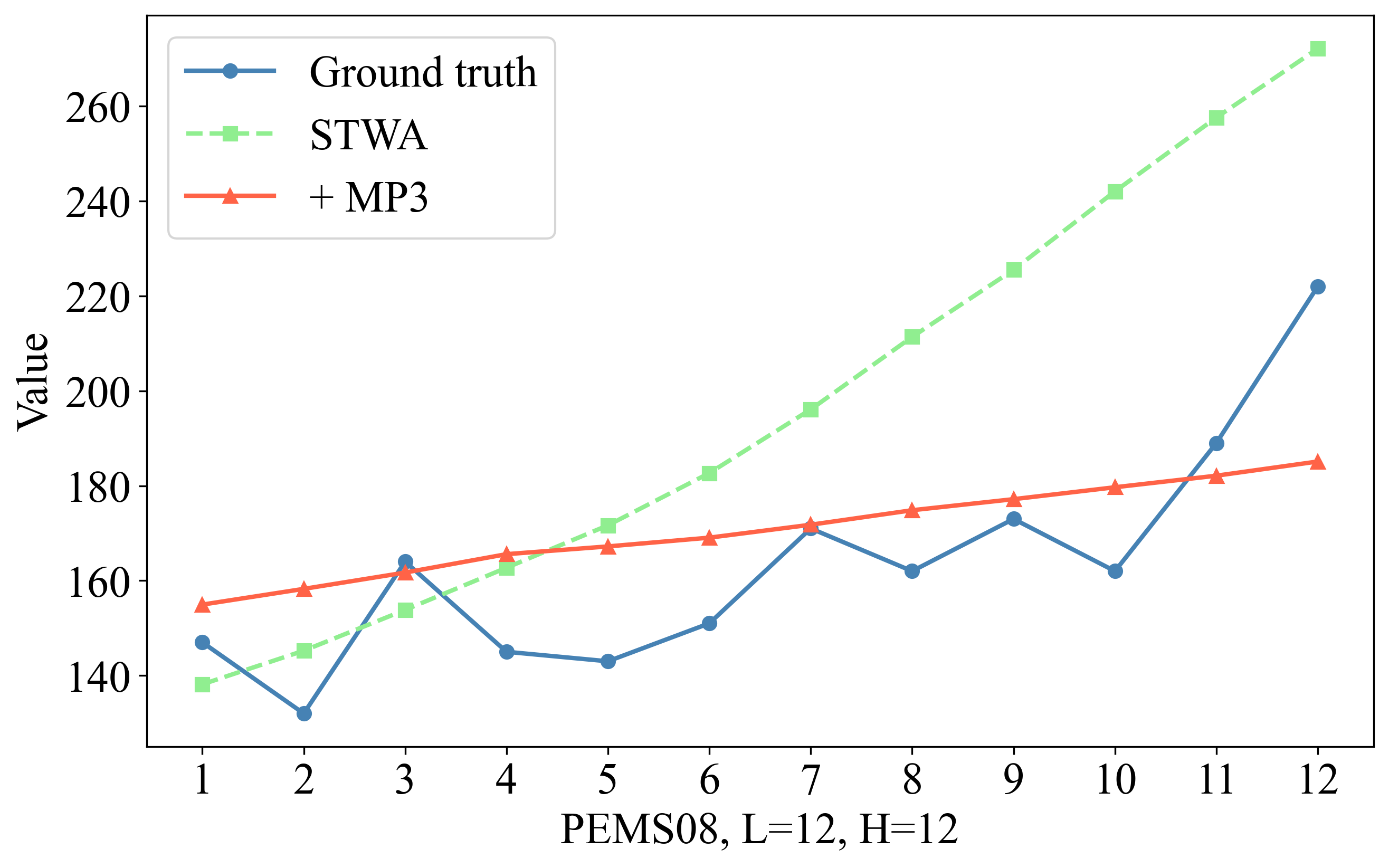}
        \label{fig:sub6}
    }

    \caption{ Figure 9 \textbf{Comparison of Forecasting Results between MP3 and Three Backbones.}}
    \label{fig:six_subplots}
\end{figure*}

\begin{figure}[!htb]
 \centering
\includegraphics[width=1\linewidth]{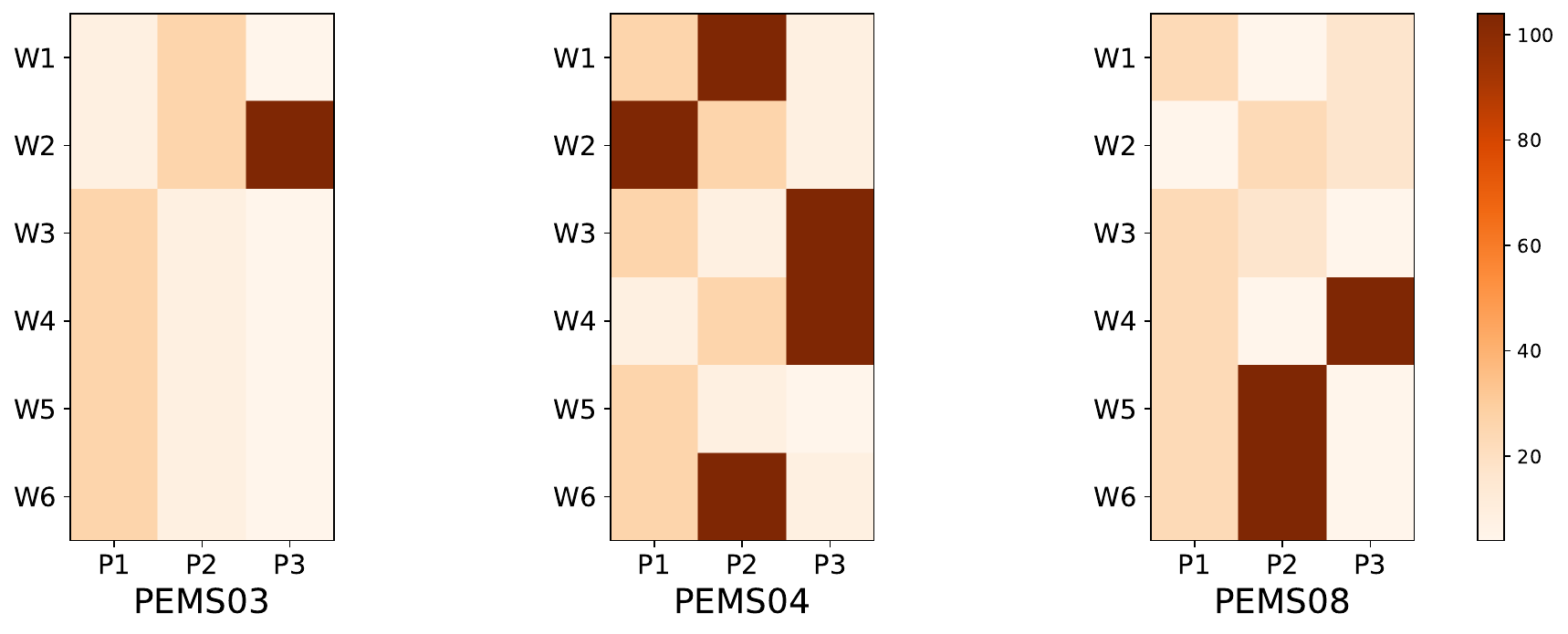} 
    \caption{Figure 10 \textbf{Dynamic Period Patterns.} }
  \label{fig11:PeriodP}
\end{figure}

\subsection{Case Study}

\subsubsection{Visualization of Forecasting Results }
To illustrate the generalizability and effectiveness of MP3 across datasets with diverse backbones, Figure 9 visualizes the prediction results of MSDR, STGCN, STWA, and their enhanced models on the PEMS04 and PEMS08 datasets.
% \ref{fig:six_subplots}. 
Based on the comparative analysis of the prediction results, the models with MP3 achieve a better fit with the ground truth, leading to more accurate prediction performance. We attribute this improvement to the proposed MP3 model, which more effectively captures long-term and dynamic Period dependencies and learn the interactions among different Period patterns. This capability enables a more precise extraction of long-term trends, which ultimately results in higher accuracy in traffic speed prediction.
\subsubsection{Dynamic multi-period patterns}
Figure 10 presents the frequency‑domain heatmaps of three significant period patterns extracted from long-time series on the PEMS03, PEMS04, and PEMS08 datasets. The figure shows that the period patterns of different input windows (w1,w2,...,w6) vary dynamically.

\begin{figure}[!htb]
 \centering
\includegraphics[width=1\linewidth]{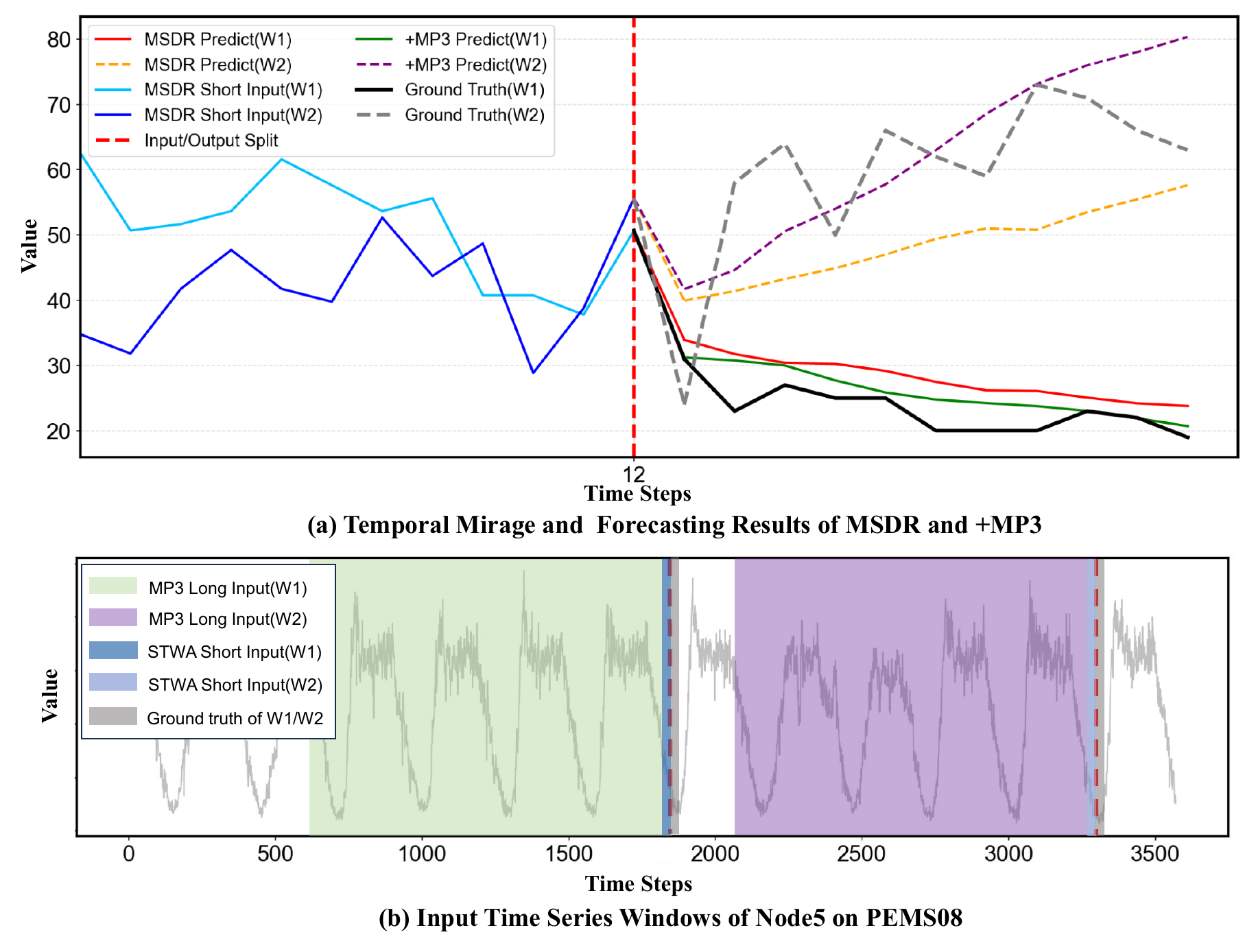} 
    \caption{Figure 11 \textbf{Temporal Mirage and Comparison of Forecasting Results}}
  \label{fig13:TM}
\end{figure}

\subsubsection{Visualization of Temporal Mirage}
Figure 11(b) illustrates input time series windows with different temporal lengths. Specifically, MP3 adopts long input windows that span a broader temporal range, enabling the capture of rich temporal context, including period patterns and long-term dependencies. In contrast, short input windows (e.g., STWA) can only access limited recent observations.

Figure 11(a) presents the corresponding forecasting results. It can be observed that two similar short input sequences correspond to entirely different future ground-truth trends, which is a typical manifestation of the temporal mirage phenomenon. Models that overly rely on short-term input patterns (e.g., MSDR) struggle to generalize when the data distribution shifts, resulting in significant performance degradation. In contrast, the predictions generated by MP3 closely align with the ground truth, demonstrating superior forecasting performance.
These results indicate that learning multi-period patterns from long input windows is crucial for enhancing model stability, mitigating the temporal mirage effect, and improving prediction accuracy.

\subsubsection{Time 2D Structure} \label{2Dstructure}
Figure 12 illustrates the Visualization of 2D temporal structures from multi-period temporal modeling. Three dominant frequencies (8, 26, 104) on PEMS04 dataset are adaptively extracted via FFT amplitude spectra. Regular block-like striped patterns are observed without row-column coupling distortions, reflecting inherent independence between intra-period and inter-period variations and supporting our decoupled edge convolution design.

\begin{figure}[!htb]
 \centering
\includegraphics[width=0.9\linewidth]{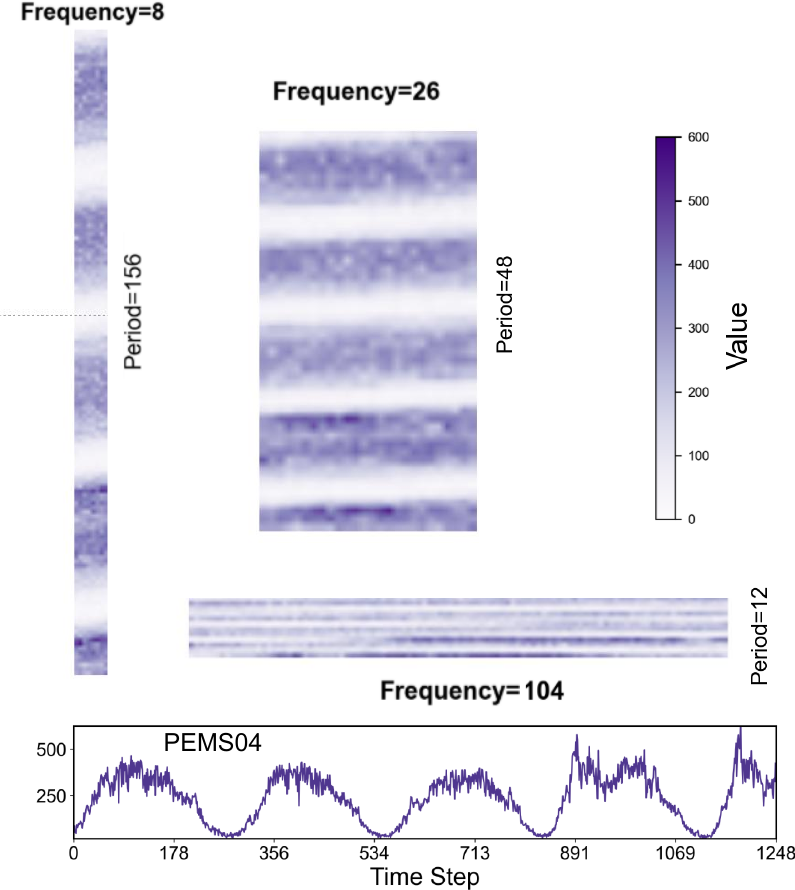} 
    \caption{Figure 12 \textbf{Case Study of 2D Temporal Structure}}
  \label{fig14:TS}
\end{figure}

\section{CONCLUSION}
This paper presented MP3, a novel plug-and-play component designed to enhance urban spatio-temporal prediction by effectively modeling multi-period spatio-temporal dependencies from long-term time series. MP3 was composed of two stages. In the pre-training stage, it distilled knowledge of multi-period patterns through multi-period temporal and spatial modeling, along with cross-period pattern interactions. In the downstream forecasting stage, MP3 was seamlessly integrated into existing spatio-temporal graph neural network backbones, strengthening their capacity to identify temporal mirage. Extensive experiments on five spatio-temporal datasets validated that MP3 consistently outperformed existing methods, confirming its effectiveness, scalability, and adaptability for spatio-temporal prediction. For future work, we plan to explore two main directions. First, we intend to integrate more rigorous causal discovery methods into cross-period interaction modeling to improve interpretability. Second, we will further enhance MP3's ability to support large-scale, low-cost inference for real-world deployment.

\bibliographystyle{elsarticle-num}
\bibliography{ref}

\end{document}